%% file: PaperForReview.tex
\documentclass[10pt,twocolumn,letterpaper]{article}

\usepackage{cvpr}              

\usepackage{graphicx}
\usepackage{amsmath}
\usepackage{amssymb}
\usepackage{booktabs}

\usepackage[pagebackref,breaklinks,colorlinks]{hyperref}

\usepackage{amsfonts}       
\usepackage{nicefrac}       
\usepackage{microtype}      
\usepackage{xcolor}         
\usepackage{amssymb}
\usepackage{amsmath}
\usepackage{breqn} 
\usepackage{pifont}
\usepackage{graphicx}
\usepackage{subcaption}
\usepackage[ruled,vlined]{algorithm2e}

\newcommand{\cmark}{\ding{51}}%
\newcommand{\xmark}{\ding{55}}%

\usepackage{tabu}

\usepackage[capitalize]{cleveref}
\crefname{section}{Sec.}{Secs.}
\Crefname{section}{Section}{Sections}
\Crefname{table}{Table}{Tables}
\crefname{table}{Tab.}{Tabs.}

\def\cvprPaperID{11110} 
\def\confName{CVPR}
\def\confYear{2022}

\begin{document}

\title{SLIC: Self-Supervised Learning with Iterative Clustering for Human Action Videos}



\author{Salar Hosseini Khorasgani*\qquad Yuxuan Chen* \qquad Florian Shkurti\\
University of Toronto \\
{\tt \small \{salar.hosseinikhorasgani, yuxuansherry.chen\}@mail.utoronto.ca, florian@cs.toronto.edu}
}





\maketitle

\begin{abstract}

Self-supervised methods have significantly closed the gap with end-to-end supervised learning for image classification~\cite{byol, chen2020simple}. In the case of human action videos, however, where both appearance and motion are significant factors of variation, this gap remains significant~\cite{coclr, xie2018rethinking}. One of the key reasons for this is that sampling pairs of similar video clips, a required step for many self-supervised contrastive learning methods, is currently done conservatively to avoid false positives. A typical assumption is that similar clips only occur temporally close within a single video, leading to insufficient examples of motion similarity. To mitigate this, we propose SLIC, a clustering-based self-supervised contrastive learning method for human action videos. Our key contribution is that we improve upon the traditional intra-video positive sampling by using iterative clustering to group similar video instances. This enables our method to leverage pseudo-labels from the cluster assignments to sample harder positives and negatives. SLIC outperforms state-of-the-art video retrieval baselines by +15.4\% on top-1 recall on UCF101 and by +5.7\% when directly transferred to HMDB51. With end-to-end finetuning for action classification, SLIC achieves 83.2\% top-1 accuracy (+0.8\%) on UCF101 and 54.5\% on HMDB51 (+1.6\%). SLIC is also competitive with the state-of-the-art in action classification after self-supervised pretraining on Kinetics400. 

\renewcommand{\thefootnote}{\fnsymbol{footnote}}
\footnotetext[1]{The two authors contributed equally to this paper.}

\end{abstract}

\section{Introduction}
\label{sec:intro}

\begin{figure}[ht]
	\centering
	
	\includegraphics[width=0.9\linewidth]{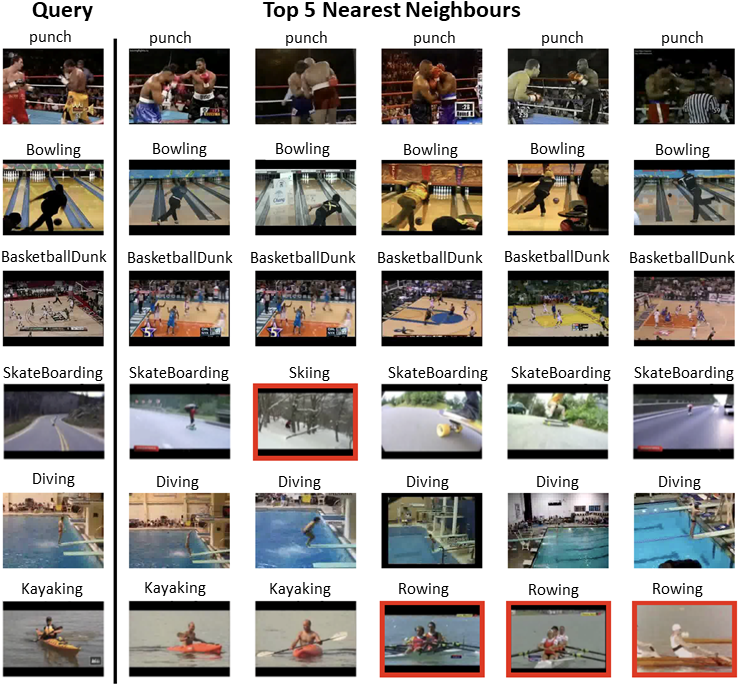}
	\caption{\textbf{Nearest neighbor retrieval results} of SLIC after pretraining on UCF101. The leftmost column is the query video from the UCF101 test set. On the right are the top 5 nearest neighbours from the training set in UCF101. Incorrect retrieval results are highlighted in red.}
	\label{fig:retrieval_results}
	\vspace{-0.3cm}
\end{figure}


Self-supervision tasks have emerged as effective pretraining methods for image classification, retrieval, and other downstream tasks. They have also been shown to outperform end-to-end supervised learning in a number of settings~\cite{supcon, chen2020simple}. A key assumption in many self-supervised methods is the ability to do instance discrimination, by sampling or generating similar and dissimilar data, given a query. While this is a reasonable assumption for training image representations, for example by generating similar data through augmentations and dissimilar data by random sampling, it becomes much more challenging in the case of video representations as we need to account for motion-based as well as appearance-based similarities.  

Different clips might have dissimilar appearances and the same motion (e.g. running at different locations), or similar appearances but different motions (e.g. playing cricket or golf with a green field as the background). 
In instance discrimination we sample positive video instances conservatively, typically by assuming that they only occur temporally close within a single video, and sample negatives randomly from different videos. 
Even though some methods attempted to improve upon instance discrimination by sampling harder positives using additional views, self-supervised pretraining of video representations is still not as effective as fully supervised pretraining (when evaluating on downstream classification using only visual inputs) ~\cite{coclr, xie2018rethinking, tran2018closer}. 

\begin{figure*}
    \vspace{0.15cm}
	\centering
	
	\includegraphics[width=0.72\textwidth]{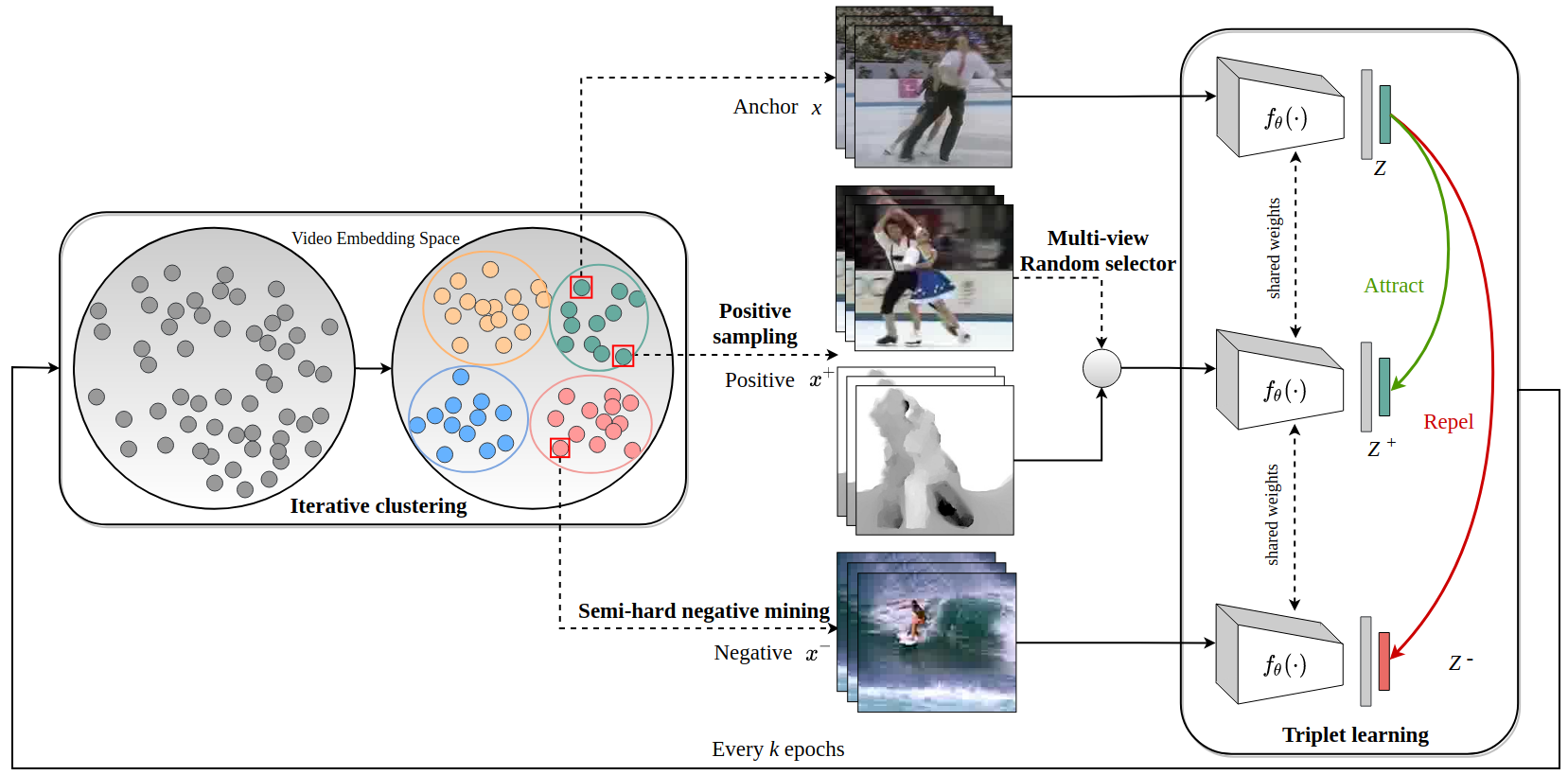}
	\caption{\textbf{Overview of the proposed self-supervised learning with iterative clustering framework (SLIC)}. We extract features using a deep 3D CNN and perform clustering every $k$ epochs in the feature space to obtain cluster assignments. The cluster assignments are used as pseudo labels to sample positives and negatives for triplet learning. There is a fixed probability of replacing the positive (RGB view) with its corresponding optical flow view. }
	\label{fig:architecture}
	\vspace{-0.5cm}
\end{figure*}

To mitigate this, we propose SLIC, a self-supervised learning method for videos. SLIC alternates between periodically clustering video representations to produce pseudo-labels, and using those pseudo-labels to inform the sampling of positive and negative pairs to update the video representations, by minimizing a triplet margin loss. SLIC also combines iterative clustering with multi-view encoding and a temporal discrimination loss to learn view-invariant embeddings and fine-grained motion features, in order to distinguish the additional aspect of similarity that arises from the temporal dimension. Figure~\ref{fig:architecture} shows an overview of our method. 

Our main contributions are twofold. First, we show that iterative clustering significantly improves upon traditional instance discrimination in self-supervised learning for video representation. While this has already been established for image representations~\cite{caron2018deep, caron2019unsupervised}, it has not been examined carefully for videos. Our method is the first to leverage efficient iterative clustering for video representation, specifically for sampling harder positives and negatives for contrastive learning. Second, we integrate iterative clustering with multi-view encoding and a temporal discrimination loss to sample harder positives and negatives during pretraining. We demonstrate that the interaction of these components, which has not been carefully examined by previous methods, is beneficial.

Our experiments show that SLIC achieves state-of-the art results on video retrieval ($+15.4\%$ improvement on top-1 recall on UCF101 and $+5.7\%$ on HMDB51 respectively, as shown in Table~\ref{tbl:NN_retrieval}), and action classification when pretrained on UCF101. When finetuning end-to-end for action classification, we observe significant gains from pretraining instead of using a random initialization of weights (about $+24\%$ on top-1 accuracy on both UCF101 and HMDB51, as shown in Table~\ref{tbl:finetune_res}). We demonstrate through additional experiments that all three components (i.e. iterative clustering, multi-view encoding, and temporal discrimination loss) complement each other, and result in larger performance improvements when combined. We evaluate the individual contribution from each component in the ablation studies, and identify that iterative clustering and multi-view encoding are the main contributing factors in SLIC (shown in Table~\ref{tbl:ablation_table}). 





\section{Related work}
\vspace{-0.1cm}

\label{related work}
\textbf{Pretext Tasks and Losses for Self-Supervision}:
Several pretext tasks have been used in the past to provide training signals for image-based self-supervision, including predicting the position of an image patch relative to the first~\cite{doersch2015unsupervised}, solving jigsaw puzzles with shuffled image patches~\cite{Noroozi2016UnsupervisedLO}, performing colorization~\cite{zhang2016colorful, vondrick2018tracking, 8237488}, inpainting of missing pixels~\cite{pathakCVPR16context}, or predicting image rotations~\cite{gidaris2018unsupervised}. In the case of video data, these tasks can be extended to include predicting the direction of time in the shuffled frames of a video~\cite{wei2018learning}, and predicting the playback rate of the video~\cite{yao2020video, Benaim_2020_CVPR, Wang20}. Our proposed method does not rely on any of these pretext tasks. 

Many of the tasks mentioned above were superseded by contrastive learning approaches, using the triplet loss~\cite{Schroff_2015}, multiclass N-pairs loss~\cite{npairs}, and variants of noise contrastive estimation~\cite{pmlr-v9-gutmann10a, oord2019representation, SeCo2021, videoMoCo, CoCon2021, VideoDeepInfoMax, DPC}, where the task is to discriminate between noisy and observed data. In particular, \cite{CoCon2021, DPC} attempt to recurrently predict the representation of future frames using previous frames, then train the network to contrast the predicted representation against the ground truth representation and a pool of distractors. Also, \cite{TimeContrastiveNetworks} uses a triplet loss to attract images from multiple videos concurrently recording different viewpoints of the same observation. In addition, ~\cite{byol} shows that negatives are not needed to achieve comparable performance with self-supervised contrastive methods in image classification. While we experimented with contrastive losses~\cite{oord2019representation, coclr, chen2020simple, chen2020mocov2}, we observed lower performance on video retrieval accuracy with iterative clustering, so our method relies on two triplet losses instead.               

\textbf{Clustering-based Self-Supervision}: Using clustering as part of the pretraining process in self-supervision has been examined in multiple prior works. For example, ClusterFit~\cite{yan2019clusterfit} clusters the features computed by a pretrained CNN using K-means and then trains a new network using the resulting cluster assignments as pseudo-labels, showing notable improvements in transfer tasks. Our method differs in two ways: (a) we do not rely on a pretrained CNN to obtain features for clustering and instead use iterative clustering steps, which Fig.~\ref{fig:training_nmi_curves} shows to reduce false positives, and (b) we do not rely on K-means and thus we do not need to pre-specify K. 

DeepCluster~\cite{caron2018deep, caron2019unsupervised} showed that the idea of using iterative clustering through K-means brings significant improvements in large-scale image classification, while our work addresses self-supervised activity recognition and video retrieval, instead. Our idea is also similar to Prototypical Contrastive Learning (PCL)~\cite{PCL}, which learns cluster prototypes and optimizes the InfoNCE loss~\cite{oord2019representation} in an Expectation Maximization~\cite{EM_algo} loop.  Though our method can be seen as an EM loop, PCL differs from our work in that it addresses image classification only, and uses a different loss for contrastive learning than what we use here. It is worth mentioning here that~\cite{hsu2016neural} presents a neural clustering method that does not compute cluster centers and thus does not require a pre-defined distance metric or number of clusters. Similarly,~\cite{YM.2020Self-labelling, swav} optimize clustering using an optimal transport solver, and a contrastive learning objective, but only address image classification and require a known number of labels, while our method does not.
Our method also shares similarities with~\cite{face_finch} which clusters face representations, however there is again the difference that we do not cluster the features from any pretrained networks and we progressively update the features used for clustering. 


\textbf{Multimodal and Multiview Self-Supervision}: Multiple modalities for self-supervision have been used as different views of the data. For example audio and video~\cite{alwassel2020selfsupervised, arandjelovic2017look, korbar, label_multimodal, cross_modal, spatio_tempoal_statistics}, or video, audio and text~\cite{akbari2021vatt}.
In particular, ~\cite{label_multimodal, cross_modal} rely on both audio and visual inputs to produce clustering assignments for cross-entropy classification (pre-specified number of clusters). 
In contrast, our method only uses visual inputs. Alternatively, ~\cite{CLIP} uses image captions as supervision to pretrain a visual model for image representation learning, but it requires additional labels for training data.

Co-training based on RGB videos and optical flow was examined in CoCLR~\cite{coclr}, which is one of the best-performing methods for video representation. \cite{CoCon2021} also adopts a multi-view training scheme similar to that of CoCLR\cite{coclr} by ensuring that the embeddings of different views such as flow, segmentation masks, and poses are consistent with respect to their distances between the same clips. 

\textbf{Action Recognition from Video}: The main approaches for supervised learning for video action recognition rely on 3D-CNN architectures~\cite{tran2018closer, feichtenhofer2019slowfast, 3d_conv_spatiotemporal} that assume either single-stream networks ~\cite{Feichtenhofer_2020_CVPR, tran2018closer,xie2018rethinking}
or multi-stream networks (i.e. separate streams for RGB and optical flow inputs)~\cite{dynamonet, simonyan, feichtenhofer2016convolutional, Zhao_2019_CVPR, feichtenhofer2019slowfast}. Our method belongs in the former category since we use one shared encoder network to process both RGB and optical flow inputs.


\section{Method}
\vspace{-0.1cm}
\label{method}
Our goal is to learn feature representations from videos in a self-supervised manner. We propose an iterative clustering based method for video representation learning. The core elements of the proposed framework consist of the following: i) performing gradient updates using a triplet margin loss, and (ii) acquiring pseudo-labels from the cluster assignments to sample triplets. 
We incorporate two loss functions to optimize the encoder: an instance-based triplet loss and a temporal discrimination loss. Furthermore, we encourage the clustering to discover motion-based similarities by incorporating multi-view encoding of the RGB and optical flow views (using a single encoder). The pseudo code is presented in Algorithm \ref{alg:training} in Appendix Section \ref{algorithm_overview}.

\subsection{Iterative Clustering}
\vspace{-0.1cm}
To generate pseudo labels for the training set, we adopt the FINCH~\cite{finch} algorithm to obtain pseudo-labels from clustering the video embeddings. FINCH discovers groupings in the data by linking the first neighbor relations of each sample, and hence does not require any prior knowledge of the data distribution. FINCH is more suitable for our task compared to other clustering methods such as K-means and DBSCAN \cite{dbscan} as it does not involve any hyper-parameter tuning or prior specification of the number of clusters, and it is considerably faster when handling large datasets.

FINCH computes the first neighbor $k^1_i$ for each video instance $x_i$ in the feature space using the cosine distance metric. For instance, $k^1_i = j$ means that $x_j$ is the first neighbor of $x_i$. It then generates an adjacency link matrix $A(i,j)$ via Equation \ref{eqn:finch}.
\begin{equation}
    \label{eqn:finch}
         A(i,j) = 
        \left\{\begin{matrix}
        1, \text{if $j=\kappa^1_i$ or $k_j^1 = i$ or $\kappa^1_i = \kappa^1_j$}
        \\
        0, \text{otherwise}
        \end{matrix}\right. 
\end{equation}

The adjacency matrix links each video instance $x_i$ to its first neighbors via $j=\kappa_i^1$, enforces symmetry through $i=\kappa_j^1$, and links video instances that share a common first neighbor with $\kappa_i^1 = \kappa_j^1$. Clustering is performed by recursively merging the connected components obtained from the adjacency matrix $A(i,j)$ in a hierarchical fashion. The output of the FINCH clustering algorithm is a small hierarchy of partitions that capture groupings in the underlying data structure at different levels of granularity, where each successive partition is a superset of the preceding partitions. Partition 1 is a flat partition of data generated via Equation \ref{eqn:finch}, which consists of a large quantity of small clusters at high purity. Due to it being the partition with the highest purity and to reduce the likelihood of sampling false positives, we use the cluster labels from the first partition to provide pseudo-labels for positive and negative mining.      

SLIC periodically performs FINCH clustering in the feature space during training every $k$ epochs. After applying FINCH, we update the pseudo label set $\{\hat{y_i}\}$ from the first partition $P_1$, where $\hat{y_i}\in \{1, 2, ..., C_{P_1}\}$, and $C_{P_1}$ is the total number of clusters generated in partition $P_1$. As an alternative to FINCH, we also experiment with K-means and spherical K-means as baselines (with different values of K). In Section \ref{section_ablations}, we demonstrate that when a large enough number of clusters is used, K-means and spherical K-means result in comparable performance to that of FINCH but at a higher computational cost. We quantify the clustering quality by computing the Normalized Mutual Information (NMI) between the pseudo labels generated by the clustering algorithm \(\{\hat{y_i}\}\) and the ground truth labels \(\{y_i\}\). 
\vspace*{-0.15cm}
\begin{equation}
    \label{eqn:nmi}
    NMI(\{\hat{y_i}\}, \{y_i\}) = \frac{I(\{\hat{y_i}\}, \{y_i\})}{\sqrt{H(\{\hat{y_i}\})H(\{y_i\})}}
\end{equation}
where $I(\cdot, \cdot)$ and $H(\cdot)$ are the mutual information and the entropy respectively. We show in Section \ref{cluster_quality} that clustering quality improves throughout the training. 

\subsection{Instance-based Triplet Loss}
\vspace{-0.1cm}
We adopt the triplet margin loss and use it in conjunction with the pseudo labels attained from clustering, which allow us to sample harder positives and negatives during training.
Let the dataset be denoted as $D$ with $N$ videos, i.e. $D=\{x_1, x_2, ..., x_N\}$, the goal is to learn an encoder $f_\theta(\cdot)$ that enforces the distance between two similar video clips to be closer than the distance between a dissimilar pair in the feature space.
Given a triplet that consists of an anchor $x$, positive $x^+$, and negative $x^-$, we want to compute $f_\theta(\cdot)$ to maximize the distance between $x$ and $x^-$ while minimizing the distance between $x$ and $x^+$, i.e. $d(f_\theta(x), f_\theta(x^{-}))> d(f_\theta(x), f_\theta(x^+))$, where $d(\cdot)$ is the cosine distance $d(f_1, f_2) = 1 - \frac{f_1 \cdot f_2 }{\|f_1\| \|f_2\|}$. The triplet margin loss is defined as follows:
\vspace{-0.1cm}
\begin{multline}
\label{eqn:triplet_loss}
     \mathcal{L}_{triplet}(x, x^+, x^-; \theta, m_1) =\\
     \text{max}(0, d(f_\theta(x), f_\theta(x^+)) - d(f_\theta(x), f_\theta(x^-)) + m_1) 
\end{multline}

\noindent where $m_1$ is a hyper-parameter that enforces a margin between anchor-positive and anchor-negative distance pairs. 

\textbf{Positive mining.} Let $x_i$ and $x_j$ be two different instances sampled from the dataset $D$, where $x_i$ is the anchor, and $x_j$ belongs to the same cluster as $x_i$. Many prior works \cite{qian2021spatiotemporal, dave2021tclr, iic} sample the positive and anchor as two non-overlapping clips from the same video, which is known as instance discrimination. Although augmented differently, these two clips have limited diversity in objects and scenes since they are sampled from the same video. Thus, it is desirable to use different videos from the same semantic class as positives. However, due to the instability of clustering quality in the early stages of training, we might sample false positives (i.e. positives sampled from the same cluster but belonging to different semantic classes) that could be detrimental to the training. Hence, we leverage both methods by sampling positives from the same instance with probability $p_\alpha$ and from different instances with probability $(1-p_\alpha)$:
\begin{equation}
    \label{eqn:pos_mining}
    x^+ = \alpha x_i^+ + (1-\alpha)x_j^+, \; \; \; \alpha \sim  \text{Bernoulli}(p_\alpha)
    \vspace{-0.1cm}
\end{equation}
where $x_i^+$ is a sampled augmented clip from the same video instance as the anchor $x_i$, and $x_j^+$ is a sampled augmented clip from a different instance, $x_j$, which belongs to the same cluster as $x_i$. $p_\alpha$ is a hyper-parameter that dictates how often the positives are sampled from the same instance or from different instances within the same cluster.

\textbf{Multi-view positives.} We denote video data from two different views as $x$ and $\texttt{view}(x)$. For our method, optical flow is used as the second view in addition to RGB to encourage learning motion-based features. We use a shared encoder to process both input views to achieve less memory consumption. To learn view-invariant embeddings, the features $f_\theta(x)$ and $f_\theta(\texttt{view}(x^+))$ should be close to each other in embedding space, since the features are extracted from similar videos. Thus, we sample the original RGB clip $x^+$ as the positive with probability $p_\beta$, or replace the RGB clip with another $\texttt{view}(x^+)$ with probability $(1-p_\beta)$. This is written as:
\vspace*{-0.2cm}
\begin{equation}
    \label{eqn:multiview}
    x^+ = \beta x^+ + (1-\beta) \texttt{view} (x^+),\; \; \;  \beta \sim \text{Bernoulli}(p_\beta)
\end{equation}

\textbf{Negative mining.} Most prior works~\cite{coclr, dave2021tclr, qian2021spatiotemporal, iic} randomly select a large batch of negatives for contrastive instance discrimination, where the easiest 95\% of the negatives do not contribute to the training and the hardest 0.1\% are usually in-class negatives~\cite{cai2020negatives}. Easy negatives are further apart from the anchor than positives, which would evaluate to a loss of zero, whereas the in-class negatives refer to the negatives that belong to the same semantic class as the anchor, and could be detrimental to the training. On the other hand, hard negatives are closer to the anchor than the positive and satisfy $d(f_\theta(x), f_\theta(x^-)) < d(f_\theta(x), f_\theta(x^+))$. To avoid easy negatives and in-class negatives, we use pseudo labels generated by the cluster assignments to sample semi-hard negatives (within the mini-batch) that satisfy the constraint:
\begin{equation}
    \label{eqn:semi_hard_neg}
    d(f_\theta(x), f_\theta(x^-)) \leq d(f_\theta(x), f_\theta(x^+)) + m_1
\end{equation}
where $x^-$ belongs to a different cluster than $x$ and $x^+$.

\subsection{Temporal Discrimination Loss}
\vspace{-0.1cm}

In order to distinguish fine-grained temporal actions, we propose a temporal discrimination loss. This loss is similar to the local-local contrastive loss proposed by TCLR~\cite{dave2021tclr}, but has the key difference that a much greater variety of negatives can be sampled across different video instances as a result of the pseudo-labels from clustering. Given an anchor clip $x$, the positive clip is designated as the spatial augmentation \(\texttt{aug}(x)\) of the anchor clip. The negative is any temporally non-overlapping clip from the same instance or, unlike TCLR~\cite{dave2021tclr}, from a different instance in the same cluster. The temporal discrimination loss is defined as:
\begin{equation}
    \label{eqn:temporal}
    \mathcal{L}_{temporal} = \mathcal{L}_{triplet}(x, \texttt{aug}(x), x^+; \theta, m_2) 
\end{equation}
where $m_2$ represents the margin parameter for the temporal discrimination loss. It should be noted here that this loss pushes away $x^+$ from $x$ whereas the instance-based triplet loss pulled it closer. To ensure that the difference between the anchor-positive and anchor-negative distances in this loss are smaller than those in the instance-based loss, $m_2$ is set to be less than $m_1$.
Overall, the combined loss function is:
\begin{equation}
    \label{eqn:total}
    \mathcal{L}_{total} = \mathcal{L}_{triplet} + \lambda \mathcal{L}_{temporal}
\end{equation}
where $\lambda$ is a weighting parameter to balance the triplet loss and the temporal discrimination loss.



\section{Experiments}

\label{experiment}
\subsection{Setup and Settings}
\label{impl_details}

\textbf{Datasets.} We conduct various experiments to evaluate the proposed model in terms of transfer and generalization. We use two video action recognition datasets for pretraining the model: UCF101 \cite{soomro2012ucf101} and Kinetics400 \cite{kay2017kinetics}. 
We use UCF101 \cite{soomro2012ucf101} and HMDB51 \cite{hmdb} for finetuning on downstream evaluation tasks. The UCF101 dataset consists of 13,320 videos spanning 101 classes, HMDB51 consists of 7000 videos in 51 action categories, and Kinetics400 consists of 222,454 training videos in 400 action classes. For self-supervised pretraining, we use the UCF101 training set (split-1) and Kinetics400 training set without any class labels. For downstream evaluation tasks, we benchmark on split-1 of both UCF101 and HMDB51.

\textbf{Implementation details.} Unless stated otherwise, we use R3D-18 \cite{hara2018spatiotemporal} as the feature extractor for all experiments due to its common use in the literature. As done in SimCLR \cite{chen2020simple}, we attach a non-linear projection head (2-layer MLP) during training, with a hidden size of 2048 and an output dimension of 128. We maintain the projection head when evaluating action retrieval, and remove it for action classification. For both training and evaluation, we use clips of 16-frames or 32-frames (consecutive frames from 25fps videos) with a spatial resolution of 128 x 128 pixels as inputs. We apply clip-wise consistent spatial augmentations including random cropping, random scaling, horizontal flipping, color jittering, color dropping, and Gaussian blurring. We also apply random temporal cropping that crops the input video clips at random time stamps. This helps the model leverage the natural variation of the temporal dimension. For multi-view positive sampling, optical flow ($u$) is computed using the unsupervised TV-L1 algorithm \cite{zach2007tvl1}, with the same pre-processing procedure as in \cite{carreira2017i3d}. We duplicate the data from the second view to generate input clips with channel dimension 3. We provide more details on the experimental setup in Appendix Section \ref{appendix_impl_details}.



\subsection{Evaluation on Video Retrieval}
\vspace{-0.1cm}

\label{video_retrieval}
In this protocol, we directly evaluate the extracted 128-dimensional representations from the self-supervised pretraining by performing nearest neighbour (NN) retrieval without any supervised fine-tuning. We follow the protocol used in prior work \cite{dave2021tclr,coclr,VCP,VCOP,Han20}, where videos from the test set are used as the query for retrieving the $k$-nearest neighbours from the training set. For every query video, we average the output representations over 10 uniformly spaced clips. For every training video, we use the representation from a clip at a random time step. For both query and training videos, spatial center-crops of the frames are used. The retrieval results are presented for both UCF101 and HMDB51 in Table \ref{tbl:NN_retrieval}. In section \ref{retrieval_results_wo_projection_head} of the Appendix, we also provide the retrieval results from discarding the projection head and using the intermediate embeddings, which attains similar performance on HMDB51 and UCF101.

\textbf{Comparison with state-of-the-art.} Similar to prior methods, we evaluate video retrieval using the top-$k$ retrieval accuracy (recall@$k$), which is defined as the accuracy that at least one of the $k$-nearest neighbours from the training set belongs to the same semantic category as the query video. As shown in Table \ref{tbl:NN_retrieval}, our proposed method achieves 71.6\% top-1 accuracy on UCF101, which is 15.4\% higher than the current state-of-the-art. The direct transferability of the trained model is evaluated on the HMDB51 dataset. SLIC achieves 28.9\% top-1 retrieval accuracy, which is 5.7\% higher than the current state-of-the-art performance from a single-stream network. 
As a reference, we also report the performance from a supervised version of SLIC which does not use iterative clustering and instead samples positives and negatives using semantic labels. The gap with this supervised model has been reduced to 9.4\% for R@1 on UCF101 and 4.1\% for R@1 on HMDB51 (using 32 frames as the temporal input size). Lastly, we include the results obtained from using S3D \cite{xie2018rethinking} as the feature extractor to demonstrate that SLIC generalizes to both ResNet and Inception-based backbones. 

\begin{table*}[ht]
    \vspace{0.2cm}

  \caption{\textbf{Nearest neighbour video retrieval results} on UCF101 and HMDB51 (both split-1). Testing set clips are used as queries to retrieve the top-$k$ nearest neighbors in the training set, where $k \in [1,5,10,20]$. \color{gray}{Models with \textsuperscript{\textdagger} use both RGB and optical flow as inputs for retrieval.} \color{black}{"Supervised SLIC" is trained using a supervised triplet loss with positives and negatives sampled using semantic labels.}} 
  \label{tbl:NN_retrieval}
  \centering

  \scalebox{0.85}{
  \begin{tabu}{lcccccc | cccc }
    \toprule
    \multicolumn{3}{c}{}    &
    \multicolumn{4}{c}{UCF101} &   
    \multicolumn{4}{c}{HMDB51}\\
    \cmidrule(r){4-11}

    Method & Input Size & Arch. & R@1 & R@5 & R@10 & R@20 & R@1 & R@5 & R@10 & R@20 \\
    \midrule
    VCP\cite{VCP}   & $16 \times 112^2$ & R3D-18 & 18.6 & 33.6 & 42.5 & 53.5 &  7.6 & 24.4 & 36.3 & 53.6 \\
    VCOP\cite{VCOP}  & $16 \times 112^2$ & R3D-18 & 14.1 & 30.3 & 40.4 & 51.1 &  7.6 & 22.9 & 34.4 & 48.8 \\
    IIC\cite{iic}   & $16 \times 112^2$ & R3D-18 & 36.5 & 54.1 & 62.9 & 72.4 & 13.4 & 32.7 & 46.7 & 61.5 \\
    \rowfont{\color{gray}}
    IIC\textsuperscript{\textdagger}  \cite{iic}  & $16 \times 112^2$ & R3D-18 & 42.4 & 60.9 & 69.2 & 77.1 & 19.7 & 42.9 & 57.1 & 70.6 \\
    MemDPC\cite{Han20}  & $40 \times 128^2$ & R3D-18 & 20.2 & 40.4 & 52.4 & 64.7 & 7.7 & 25.7 & 40.6 & 57.7 \\
    CoCLR-RGB\cite{coclr}  & $32 \times 128^2$ & S3D-23 & 53.3 & 69.4 & 76.6 & 82.0 & 23.2 & 43.2 & 53.5 & 65.5 \\
    \rowfont{\color{gray}}
    CoCLR\textsuperscript{\textdagger}\cite{coclr}  & $32 \times 128^2$ & S3D-23 & 55.9 & 70.8 & 76.9 & 82.5 & 26.1 & 45.8 & 57.9 & 69.7 \\
    TCLR\cite{dave2021tclr}    & $16 \times 112^2$ & R3D-18 & 56.2 & 72.2 & 79.0 & 85.3 & 22.8 & 45.4 & 57.8 & 73.1 \\
    \midrule
    
    SLIC & $16 \times 128^2$ & S3D-23 & 60.9 & 73.5 & 78.6 & 84.0 & 21.8 & 46.3 & 59.7 & 72.4 \\
    SLIC & $32 \times 128^2$ & S3D-23 & 69.8 & 79.2 & 83.2 & 87.2 & 26.8 & \textbf{52.9} & \textbf{66.2} & \textbf{78.1} \\
    SLIC & $16 \times 128^2$ & R3D-18 & 66.7 & 77.3 & 82.0 & 86.4 & 25.3  & 49.8 & 64.9 & 76.1 \\
    SLIC & $32 \times 128^2$ & R3D-18 & \textbf{71.6} & \textbf{82.4} & \textbf{86.6} & \textbf{90.3} & \textbf{28.9} & 52.8 & 65.4 & 77.8\\
    

    \midrule
    \rowfont{\color{black}}

    Supervised SLIC & $16 \times 128^2$ & S3D-23 & 67.7 & 73.2 & 74.8 & 76.0 & 17.3 & 39.4 & 53.7 & 67.7\\
    Supervised SLIC & $32 \times 128^2$ & S3D-23 & 72.5 & 79.1 & 80.9 & 83.9 & 19.1 & 45.1 & 58.0 & 70.7\\
    
    Supervised SLIC & $16 \times 128^2$ & R3D-18 & 75.3 & 80.5 & 82.8 & 84.9 & 26.8 & 51.2 & 65.1 & 76.3 \\
    Supervised SLIC & $32 \times 128^2$ & R3D-18 & 81.0 & 84.9 & 86.5 & 88.6 & 33.0 & 57.4 & 70.0 & 82.7 \\
    
    \bottomrule
  \end{tabu}
  }
  \vspace{-0.5cm}
\end{table*}

\textbf{Qualitative results.} Figure \ref{fig:retrieval_results} visualizes 6 query videos sampled from the test set and their top 5 nearest neighbors from the training set for UCF101. SLIC is able to retrieve similar videos based on their high-level semantics, and we observe that the incorrect retrievals generally share similar motion patterns and scene dynamics with the query video.

\subsection{Evaluation on Action Classification}
\vspace{-0.1cm}

We pretrain the model on UCF101 and Kinetics 400, then evaluate the pretrained model for action recognition on UCF101 and HMDB51 following the protocol from prior work \cite{coclr}. After self-supervised pretraining, we discard the non-linear projection head and attach a linear classifier, then evaluate the model under two settings: i) linear probing with a frozen backbone, and ii) fine-tuning the network end-to-end. The model is trained using the supervised cross-entropy loss, while applying all the data augmentations mentioned in Section \ref{impl_details} except for Gaussian blur. We use a learning rate of $10^{-3}$ for the linear classifier, and $10^{-4}$ learning rate for the backbone (when not frozen). For both i) and ii), the learning rate is decayed by a factor of 0.1 at epoch 60 and 100, and the training is carried out for 150 epochs. During inference, we follow the same procedure as \cite{coclr, Han20} and average the probabilities from 10 spatial crops (center-crop plus four corners, with horizontal flipping), and all temporal crops from a moving window with half temporal overlap. 

\textbf{Comparison with state-of-the-art.} Comparison with the results from prior works are presented in Table \ref{tbl:finetune_res}. When pretrained on UCF101, SLIC outperforms the current state-of-the-art on end-to-end finetuning and linear probing using a temporal window of 32 frames. Compared to our fine-tuning results from using randomly initialized weights, self-supervised pretraining on UCF101 provides significant improvement in classification accuracy for both UCF101 and HMDB51. Furthermore, to examine how action classification performance would vary if semantic label information was used during pretraining, we perform end-to-end finetuning and linear probing on Supervised SLIC. Compared to SLIC, Supervised SLIC pretraining improves linear probing accuracy by 2-5\% (depending on temporal input size and dataset) on UCF101 and HMDB51, but only provides marginal improvements for end-to-end finetuning. 
When pretrained on the larger Kinetics400 dataset, SLIC achieves comparable performance to TCLR\cite{dave2021tclr} but underperforms compared to CoCLR-RGB\cite{coclr}. 


\begin{table}[h!]
  \caption{\textbf{Linear probing and end-to-end finetuning top-1 accuracy results for action classification} pretrained on UCF101 and Kinetics400, then fine-tuned on UCF101 and HMDB51, using only visual inputs. Methods with * indicate that results were averaged across the 3 splits of each dataset instead of only using split-1. Subscripts of the input sizes indicate the skip rates each method uses.  
  "None (Rand. Init.)" is trained using a supervised cross-entropy loss with a randomly initialized backbone. 
  \color{gray}{Methods with \textsuperscript{\textdagger} use optical flow or residual frames in addition to RGB as inputs to the action classification model.}
  \color{black}{SLIC\textsuperscript{\textasciicircum} indicates that pretraining used 16 frame inputs. Dashes indicate unavailable information.}
  \color{black}{"Supervised SLIC" is trained using a supervised triplet loss with positives and negatives sampled using semantic labels. "Cross-Ent." is the state-of-the-art finetuning performance after supervised pretraining (cross-entropy loss) on Kinetics400.}}
  \label{tbl:finetune_res}
    \vspace{-0.1cm}

  \centering
  
  \resizebox{\columnwidth}{!}{%
  \begin{tabu}{lcccc}
    \toprule
    Pretrain. Method & Input Size & Arch. & UCF101 & HMDB51 \\
    \toprule
    None (Rand. Init.) & $16 \times 128^2$ & R3D-18  & 53.5 & 22.3	 \\
    \toprule
    \multicolumn{5}{c}{\textbf{Linear Probe (Pretrained on UCF101)}}\\
    \midrule
    TCLR\cite{dave2021tclr} & $16_{\times2} \times 112^2$ & R3D-18 &  69.9 & - \\
    CoCLR-RGB\cite{coclr}  & $32 \times 128^2$ & S3D-23 & 70.2 & 39.1 \\
    \rowfont{\color{gray}}
    CoCLR\textsuperscript{\textdagger}\cite{coclr} & $32 \times 128^2$ & S3D-23 & 72.1 & 40.2 \\
    \midrule
    SLIC  & $32 \times 128^2$ & S3D-23 & 74.8 & 44.7	 \\

    SLIC  & $16 \times 128^2$ & R3D-18 & 72.3 & 41.8	 \\
    \textbf{SLIC} & $32 \times 128^2$ & R3D-18 & \textbf{77.7} & \textbf{48.3} 	 \\
    \midrule
    \rowfont{\color{black}}
    Supervised SLIC & $16 \times 128^2$ & R3D-18  & 76.4 & 44.9	 \\
    Supervised SLIC & $32 \times 128^2$ & R3D-18  & 82.3 & 50.6	 \\
    \toprule
    \multicolumn{5}{c}{\textbf{End-to-end Finetuning (Pretrained on UCF101)}}\\
    \midrule
    DPC\cite{DPC} & $40\times128^2$ & R3D-18 & 60.6 & -\\
    VCP\textsuperscript{*}\cite{VCP}  & $16 \times 112^2$ & R3D-18 & 66.0 &	31.5 \\
    VCOP\textsuperscript{*}\cite{VCOP} & $16 \times 112^2$ & R3D-18 &  64.9 & 29.5 \\
    IIC\cite{iic}  &  $16 \times 112^2$ & R3D-18 &  61.6  & - \\
    \rowfont{\color{gray}}
    IIC\textsuperscript{\textdagger*}\cite{iic} & $16 \times 112^2$ & R3D-18 & 74.4  & 38.3 \\
    MemDPC\cite{Han20}  & $40 \times 128^2$ & R3D-18 &  68.2 &	- \\
    CoCLR-RGB\cite{coclr}   & $32 \times 128^2$ & S3D-23 &81.4 &	52.1 \\
    \rowfont{\color{gray}}
    CoCLR\textsuperscript{\textdagger}\cite{coclr} & $32 \times 128^2$ & S3D-23 & \textbf{87.3} &	\textbf{58.7} \\
    TCLR\textsuperscript{*}\cite{dave2021tclr} & $16_{\times2} \times 112^2$ & R3D-18 &  82.4 & 52.9 \\
    CoCon-RGB\textsuperscript{*}\cite{CoCon2021} & $(-)\times 224^{2}$ & R3D-18 & 70.5 & 38.4 \\
    \rowfont{\color{gray}}
    CoCon-E\textsuperscript{*}\textsuperscript{\textdagger}\cite{CoCon2021} & $(-)\times 224^{2}$ & R3D-18 & 82.4 & 52.0 \\
    \midrule
    SLIC  & $32 \times 128^2$ & S3D-23 & 81.3 & \textbf{56.2}	 \\

    SLIC  & $16 \times 128^2$ & R3D-18 &  77.4 & 46.2 \\
    \textbf{SLIC}  & $32 \times 128^2$ & R3D-18 &  \textbf{83.2} & 54.5  \\ 

    \midrule
    \rowfont{\color{black}}
    Supervised SLIC & $16 \times 128^2$ & R3D-18  & 77.4 & 46.6	 \\
    Supervised SLIC & $32 \times 128^2$ & R3D-18  & 81.9 & 56.1	 \\
    \toprule
    \multicolumn{5}{c}{\textbf{End-to-end Finetuning (Pretrained on Kinetics400)}}\\
    \midrule
    DPC\cite{DPC} & $40\times128^2$ & R3D-18 & 68.2 & 34.5\\
    CoCon-RGB\textsuperscript{*}\cite{CoCon2021} & $(-)\times 224^{2}$ & R3D-18 & 71.6 & 46.0 \\
    \rowfont{\color{gray}}
    CoCon-E\textsuperscript{*}\textsuperscript{\textdagger}\cite{CoCon2021} & $(-)\times 224^{2}$ & R3D-18 & 78.1 & 52.0 \\
    VideoMoCo\cite{videoMoCo} & $16\times112^2$ & R3D-18 & 74.1 & 43.6\\
    RSPNet\cite{chen2020RSPNet} & $16\times112^2$& R3D-18 & 74.3 & 41.8 \\
    SpeedNet\cite{Benaim_2020_CVPR} & $16\times224^2$ & S3D-G & 81.1 & 48.8 \\
    TCLR\textsuperscript{*}\cite{dave2021tclr} & $16_{\times2} \times 112^2$ & R3D-18 & 84.1 &	53.6 \\
    \textbf{CoCLR-RGB\cite{coclr}} & $32 \times 128^2$ & S3D-23 & \textbf{87.9} &	\textbf{54.6} \\
    \rowfont{\color{gray}}
    CoCLR\textsuperscript{\textdagger}\cite{coclr} & $32 \times 128^2$ & S3D-23 & 90.6 & 62.9 \\
    \midrule
    SLIC\textsuperscript{\textasciicircum} & $32 \times 128^2$ & S3D-23  & 82.3 & 49.4 \\
    SLIC & $32 \times 128^2$ & S3D-23  & \textbf{83.5} & 51.7  \\
    SLIC\textsuperscript{\textasciicircum} & $32 \times 128^2$ & R3D-18  & 83.2 & \textbf{52.2} \\
    SLIC & $32 \times 128^2$ & R3D-18  & 83.1 & 52.0 \\
    \midrule
    \rowfont{\color{black}}
    Cross-Ent. \cite{tran2018closer}& $16 \times 112^2$ & R(2+1)D-34 & 96.8 & 74.5	 \\
    \bottomrule

  \end{tabu}%
  }
\end{table}

\subsection{Evolution of Cluster Assignments and Retrieval Accuracy}
\vspace{-0.1cm}

\begin{figure}
\vspace{0.15cm}
\centering
\begin{subfigure}{0.155\textwidth}
  \centering
  \includegraphics[width=1.0\linewidth]{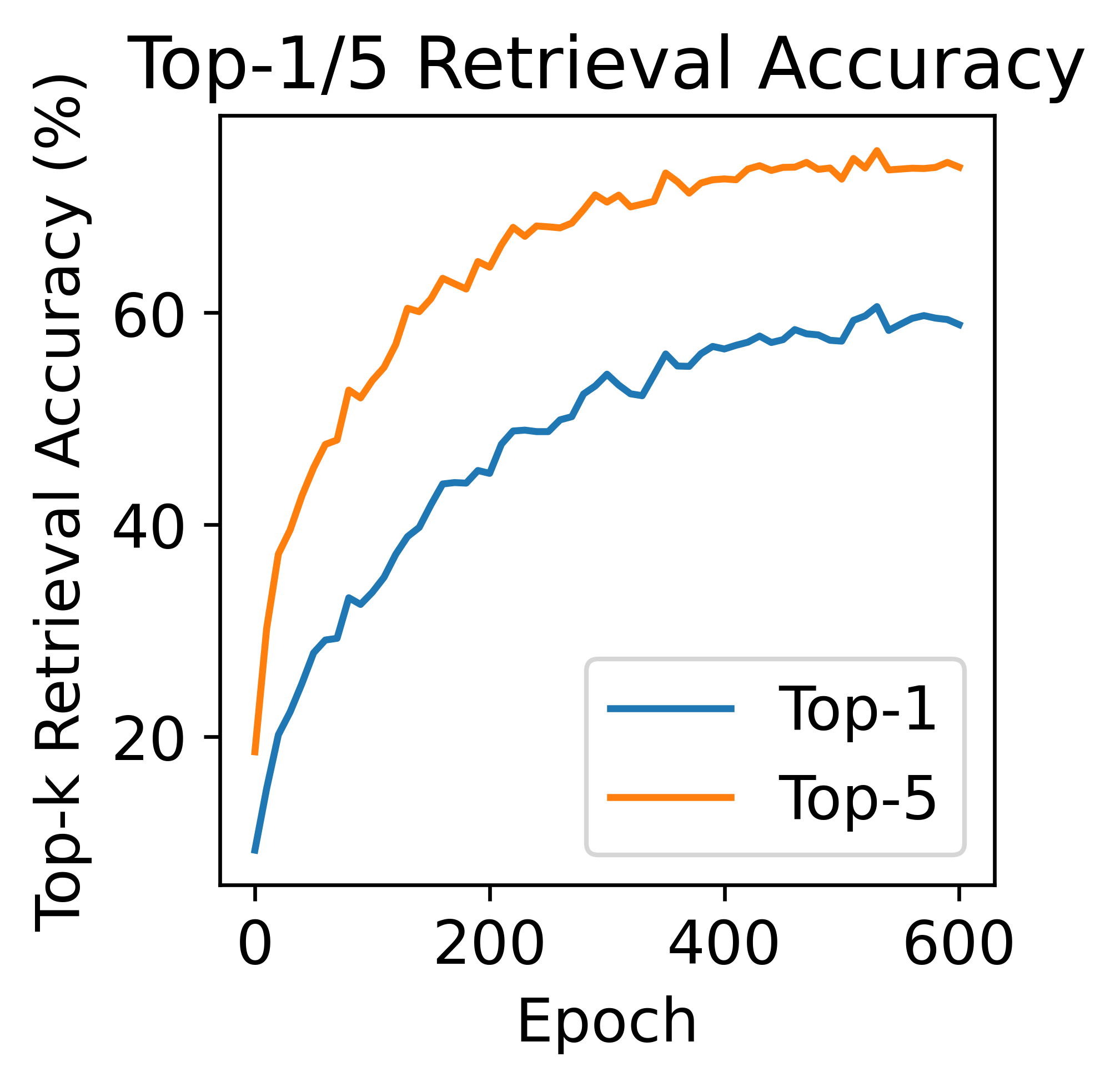}
\end{subfigure}%
\begin{subfigure}{0.155\textwidth}
  \centering
  \includegraphics[width=1.03\linewidth]{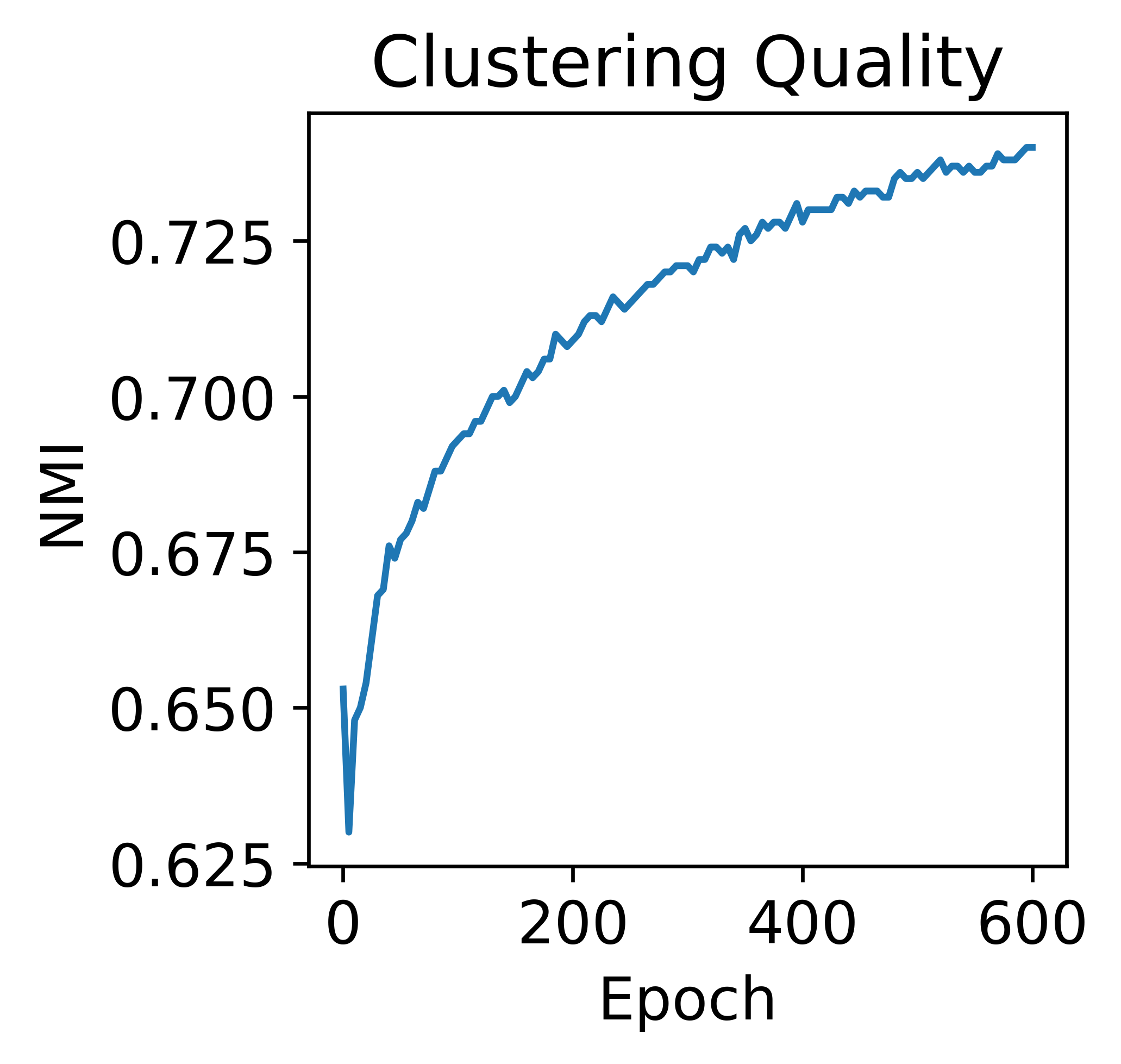}
\end{subfigure}
\begin{subfigure}{0.155\textwidth}
  \centering
  \includegraphics[width=0.97\linewidth]{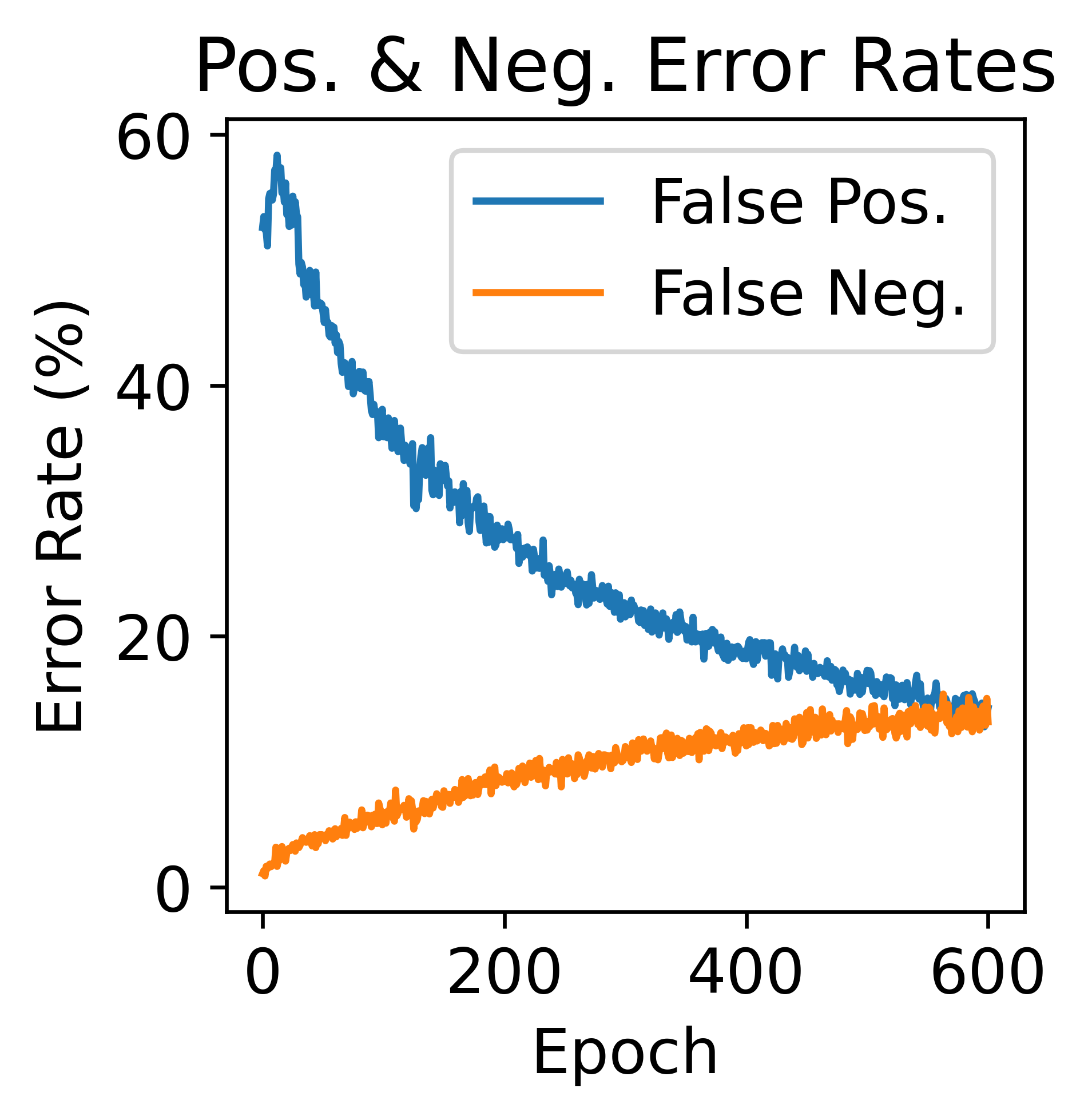}
\end{subfigure}
\vspace{-0.2cm}
\caption{\textbf{Retrieval accuracy, clustering quality, and false sampling rates} for the experiment settings described in section \ref{impl_details} over 600 epochs on UCF101. (left): evolution of the top-1/5 retrieval accuracy; (middle): evolution of the clustering quality as measured by the NMI between the cluster assignments and the ground-truth labels; (right): percentage of positives and negatives sampled from clusters which are erroneous according to ground truth labels.}
\label{fig:training_nmi_curves}
\vspace{-0.5cm}
\end{figure}

\label{cluster_quality}
This section demonstrates the evolution of the clustering quality, and a simplified top-$k$ retrieval accuracy which only uses the center 16 or 32 frames of query (test) and training videos as inputs. This simplified retrieval accuracy was used to monitor training progress since it is faster to compute than the final metric described in Section \ref{video_retrieval}. As shown in Figure \ref{fig:training_nmi_curves}, the retrieval accuracy plateaus by the end of training when the clustering quality (the NMI between the cluster assignments and the ground-truth semantic labels) has nearly stabilized. Thus, the embeddings gradually improve in their ability to represent information relating to semantic classes. Additionally, the false positive rate (percentage of positives sampled using cluster assignments which do not have the same ground-truth label as the anchor) is decreasing during training, which indicates that the sampled positives get better as the clustering quality improves. On the other hand, the false negative rate (negatives with the same ground-truth label as the anchor) increases as training progresses, indicating that some videos with the same semantic label are being grouped in different clusters. The inability to detect these false positive and negatives is one of the current limitations of our method. 


\subsection{Ablation Studies}
\vspace{-0.1cm}

\label{section_ablations}

\textbf{Ablation study on major training components.} In order to study the impact of the main training components used in SLIC, we test the top-1/5 retrieval performance of R3D-18 models pretrained on UCF101 with a subset of the components. The results from removing iterative clustering, multi-view positive sampling, and the temporal discrimination loss from the self-supervised pretraining are shown in Table \ref{tbl:ablation_table}. With all-else fixed, the addition of iterative clustering results in a significant $21.7\%$ gain (from $45.0\%$) in R@1 on UCF101 and a $5.8\%$ gain (from $19.5\%$) in R@1 on HMDB51. The next-best impact on UCF101 R@1 performance comes from mult-view positive sampling ($59.2\%$ without it). However, the largest R@1 performance gain on HMDB51 comes from the temporal discrimination loss ($19.1\%$ without it), possibly indicating that HMDB51 benefits more from temporal attention than UCF101.

\begin{table}[h!]
    \vspace{-0.2cm}

  \caption{\textbf{Ablation study on the impact of 3 major training components}: (1) iterative clustering (IC), (2) multi-view positive replacement using optical flow (MV), and (3) temporal discrimination loss (TL). }
      \vspace{-0.1cm}

  \label{tbl:ablation_table}
  \centering
  \resizebox{\columnwidth}{!}{%
  \begin{tabular}{cccccc | cc}
    \toprule
    \multicolumn{4}{c}{} & \multicolumn{2}{c}{UCF101} & \multicolumn{2}{c}{HMDB51}\\
    \cmidrule(rl){5-8}
    IC & MV & TL & Input Size & R@1 & R@5 & R@1 & R@5 \\
    \midrule
    \xmark & \cmark & \cmark & $16 \times 128^2$  & 45.0 & 62.3  & 19.5 & 45.1     \\
    \midrule
    \cmark & \xmark & \xmark & $16 \times 128^2$  &54.7 & 65.6 & 18.2 & 41.5    \\
    \cmark & \cmark  & \xmark & $16 \times 128^2$  & 59.9 & 69.8 & 19.1 & 41.1    \\
    \cmark & \xmark  & \cmark & $16 \times 128^2$ &  59.2 & 69.8  & 20.1 & 43.6    \\
    \cmark & \cmark & \cmark & $16 \times 128^2$ & \textbf{66.7} & \textbf{77.3} & 
                                 \textbf{25.3} & \textbf{49.8} \\

    \bottomrule
  \end{tabular}%
  }
\end{table}

\textbf{Ablation studies on clustering settings.} We perform ablation studies to measure the impact of the clustering settings used during the self-supervised pretraining. In particular, we try clustering with FINCH partition-2, K-means, and Spherical K-means as alternatives to FINCH partition-1, and we measure the impact of the cluster interval ($k$) as well as the number of clusters in the case of K-means. All experiments are performed with $16\times128^2$ input size. As shown in table \ref{tbl:clustering-components_finch}, K-means and Spherical K-means result in comparable performance to FINCH partition-1 (which results in approximately 2200 clusters when clustering UCF101 video embeddings) when at least 1000 clusters are used. However, FINCH is significantly more efficient than K-means (roughly 85x faster) when clustering embeddings from the larger Kinetics400 dataset. Furthermore, the performance of K-means is quite sensitive to the number of clusters. For instance, using only 100 clusters with K-means results in a significant degradation in retrieval accuracy. The hyper-parameter search would pose great challenge when deploying K-means in practice. Hence, FINCH is a more suitable option for this application since it doesn't require additional hyper-parameter tuning.
In addition, the performance of FINCH partition-2 (which results in approximately 550 clusters) is lower than that of the first FINCH partition. Thus, it can be hypothesized that over-clustering, or using many clusters with high purity, is beneficial to positive sampling. Additionally, cluster intervals of 1, 2, 5, and 10 result in similar video retrieval performance. Thus, it is not necessary to re-compute all training set embeddings and cluster them every epoch to achieve the best performance.

\begin{table}[h]
      \vspace{-0.2cm}

  \caption{\textbf{Ablation study on the impact of clustering settings} on video retrieval (after pretraining on UCF101). `Num.' indicates the number of clusters and `$k$' indicates the clustering interval. } 
      \vspace{-0.2cm}

  \label{tbl:clustering-components_finch}
  \centering
  \resizebox{\columnwidth}{!}{%
  \begin{tabular}{lcccc|cc}
  
    \toprule
    \multicolumn{3}{c}{} & \multicolumn{2}{c}{UCF101} & \multicolumn{2}{c}{HMDB51} \\
    \cmidrule(rl){4-7}
    Clustering Method & Num. & $k$ & R@1 & R@5 & R@1 & R@5 \\
    \midrule
    FINCH(partition 1) & -     & 1  & 66.8  & 77.9 & 24.5 & 49.0   \\
    FINCH(partition 1) & -     & 2  & 64.8 & 76.7 & 24.2 & 46.6    \\
    \textbf{FINCH(partition 1)}& -     & \textbf{5}  & \textbf{66.7} & \textbf{77.3} & \textbf{25.3} & \textbf{49.8}     \\
    FINCH(partition 1) & -     & 10 & 66.7 & 77.8 & 22.4 & 48.7    \\
    \midrule
    FINCH(partition 2) & -     & 5  & 62.2 & 76.1 & 21.0 & 44.7    \\
    \midrule
    K-means             & 100   & 5  & 51.5 & 66.4 & 18.7 & 43.0    \\
    K-means             & 500   & 5  & 63.5 & 76.2 & 22.0 & 47.4    \\
    K-means             & 1000  & 5  & 66.0 & 76.4  & 21.8 & 46.4     \\
    K-means             & 2000  & 5  & 66.4 & 77.3 & 23.2 & 47.9     \\
    \midrule
    Spherical K-means   & 1000  & 5  & 64.9  & 77.3  & 23.6  & 48.2     \\
    \bottomrule
  \end{tabular}%
  }
  \vspace{-0.15cm}

\end{table}

\vspace{-0.15cm}

\textbf{Ablation study on positive sampling settings.} We perform ablation studies to measure the impact of different positive sampling settings on the nearest neighbour retrieval tasks. $p_{\alpha}$ determines the probability of sampling positives from the same video instance (as opposed to different instances within the same cluster), and $p_{\beta}$ determines the probability of using RGB inputs for positives (as opposed to replacing them with the optical flow view). The retrieval results from varying $p_{\alpha}$ and $p_{\beta}$ are shown in Table \ref{tbl:positive-sample}. All experiments are performed with $16\times128^2$ input size. These results indicate that it is beneficial to sample most, but not all, of the positive clips using clustering assignments. Further, increasing $p_{\beta}$ results in improved performance. The best performance is achieved with $p_{\alpha}=0.2$ and $p_{\beta}=0.75$. 
\begin{table}[h!]
    \vspace{-0.2cm}

  \caption{\textbf{Ablation study on the impact of \boldmath$p_{\alpha}$ and \boldmath$p_{\beta}$} on video retrieval (after pretraining on UCF101). The ablations on $p_{\alpha}$ are done using RGB views only. }
      \vspace{-0.2cm}

  \label{tbl:positive-sample}
  \centering
  \resizebox{\columnwidth}{!}{%
  \begin{tabular}{lcccc|cc}
    \toprule
    \multicolumn{3}{c}{} & \multicolumn{2}{c}{UCF101} & \multicolumn{2}{c}{HMDB51}\\
    \cmidrule(rl){4-7}
    Pos. View & $p_{\alpha}$ & $p_{\beta}$ & R@1 & R@5 & R@1 & R@5 \\
    \midrule
    RGB  & 0.0 & - & 54.8 & 67.1 & 16.3 & 36.9 \\
    RGB  & 0.2 & -& 59.2 & 69.8 & 20.1 & 43.6  \\
    RGB  & 0.5  & -& 52.3 & 67.5 & 15.7 & 36.5  \\
    RGB  & 0.7 & -& 52.8 & 67.6 & 16.2 & 36.2 \\
    \midrule
    Optical flow & 0.2 & 0.25 & 59.6 & 74.3 & 21.2 & 45.7 \\
    Optical flow & 0.2 & 0.5   & 62.8 & 76.2 & 21.3 & 46.9  \\
    \textbf{Optical flow}  & \textbf{0.2} & \textbf{0.75}  & \textbf{66.7} & \textbf{77.3} & \textbf{25.3} & \textbf{49.8} \\
    \bottomrule
  \end{tabular}%
  }
  \vspace{-0.5cm}
\end{table}
\section{Limitations and Future Work}

One of the main limitations of our method is that it is sensitive to the false positive and false negative rate. This is mainly an issue for the former since it was observed in Figure \ref{fig:training_nmi_curves} that the false positive rate starts quite high at $~60\%$, but is reduced due to iterative clustering. This high false positive rate could also potentially be reduced by obtaining multiple clustering assignments (e.g., from multiple clustering algorithms or clustering embeddings from multiple input views) and prioritizing positives which are in agreement across all cluster assignment sets. 
Additionally, it was observed in Figure \ref{fig:training_nmi_curves} that the false negative rate was rising during the training, which indicates that some semantically similar videos are considered as dissimilar pairs when training the model. This could be detrimental to the final performance. We plan to address this issue by continuing to sample positives from one of the lower FINCH partitions but instead sample negatives from a higher partition. Since the higher FINCH partitions capture the data structure at a coarser granularity, it would reduce the likelihood of sampling false negatives when sampling from a higher partition. 


\vspace{-0.2cm}

\section{Conclusions and Societal Impact}
\vspace{-0.2cm}
\label{conclusion}
We propose a self-supervised, iterative clustering based contrastive learning framework to learn useful video representations. We showed that we can sample harder positives through clustering, and achieve further improvement by incorporating multi-view positives and a temporal discrimination loss. Our proposed method, SLIC, achieves state-of-the-art performance across various downstream video understanding tasks. Finally, while there is both promise and need for self-supervised representations to enable better activity recognition, video retrieval, and other downstream tasks, there exist potential uses of this technology (e.g. surveillance) that could exacerbate societal problems.     

\noindent \textbf{Acknowledgment} This work is supported by the Vector Scholarship in Artificial Intelligence, provided through the Vector Institute.

{\small

\bibliographystyle{ieee_fullname}
\bibliography{egbib}
}

\clearpage

\input{appendix}

\end{document}

%% file: appendix.tex
\appendix
\label{Appendix}

\section{Appendix overview}
The appendix is organized into the following sections:
\begin{itemize}
  \item Section \ref{appendix_impl_details}: Hyperparameter and training details
  \item Section \ref{architecture_details}: Architecture details
  \item Section \ref{algorithm_overview}: Training Algorithm Overview
  \item Section \ref{slic_infonce}: Pretraining using the InfoNCE loss
  \item Section \ref{retrieval_results_wo_projection_head}: Retrieval Results without Projection Head
  \item Section \ref{action_classif_dualview}: Classification using RGB and optical flow
  \item Section \ref{tsne_visualization}: Visualization of learned representations
  \item Section \ref{confusion_matrix_classification}: Confusion matrix for action classification
\end{itemize}

\section{Hyperparameter and training details}
\label{appendix_impl_details}
In addition to the implementation settings described in Section \ref{impl_details}, we describe here the hyperparameter and training settings used for the experiments shown in Section \ref{experiment}.

The triplet margins used are $m_1=0.2$ and $m_2=0.04$ for the instance-based and temporal discrimination triplet losses respectively, and the weighting on the temporal discrimination loss is $\lambda=1$. The embeddings that are used for clustering are computed using the temporal center-crops (16 or 32 frames) of the videos in the training set. The cluster interval $k$ is set to 5 epochs (i.e. clustering is performed for every 5 epochs during training), the probability of sampling positives from the same video (as opposed to the same cluster) is $p_\alpha=0.2$ for UCF101 and $p_\alpha=0.5$ for Kinetics400, and $p_\beta$ is set to 0.75 (the probability of using optical flow as the positive is 0.25). We observe that a higher $p_\alpha$ works better for Kinetics400 since, relative to UCF101, the increased number of classes and videos make it harder to produce high quality clusters in the beginning of the pretraining.

For optimization, the SGD optimizer was used with a fixed learning rate of 0.1, a momentum of 0.5, and no weight decay. UCF101 training was done on 2 GPUs with a batch size of 16 samples per GPU, and lasted for 600 epochs. Kinetics400 training was done on 8 GPUs with a batch size of 13 samples per GPU, and lasted for 340 epochs.

\section{Architecture details}
\label{architecture_details}
We use the 3D ResNet-18 (R3D-18) architecture \cite{hara2018spatiotemporal} for all experiments. During pretraining, R3D-18 is followed by a non-linear projection head to project the feature embeddings onto a 128-dimensional space, as done in SimCLR \cite{chen2020simple} and CoCLR \cite{coclr}. The projection head consists of two fully-connected layers, with batch normalization and ReLU activation after the first fully-connected layer. The output from the projection head is directly used for the nearest neighbour retrieval task. When evaluating on action classification, the projection head is replaced with a single linear layer. The detailed architecture is shown in Table \ref{tab:resnet_tb}.

\begin{table}[ht]
	\centering
	\caption{\textbf{3D ResNet-18 Architecture \cite{hara2018spatiotemporal}}. The projection head is used during the pretraining stage and the nearest neighbour evaluation. During finetuning, the projection head is discarded, and a single linear layer is attached after the average pooling layer.}
	\label{tab:resnet_tb}
    \resizebox{\columnwidth}{!}{%
    \begin{tabular}{ p{3cm} p{7cm} }
     \toprule
     Layer Name & Architecture \\
     \midrule
     conv1 & $7\times7\times7$, 64, stride 1 (T), 2(XY) \\
     \midrule
     conv2\_x & $3\times3\times3$ max pool, stride 2 \\
      & $\begin{bmatrix}
            3\times3\times3, & 64\\ 
            3\times3\times3, & 64 
            \end{bmatrix} \times 2$\\
     \midrule
     conv3\_x & $3\times3\times3$ max pool, stride 2 \\
      & $\begin{bmatrix}
            3\times3\times3, & 128\\ 
            3\times3\times3, & 128 
            \end{bmatrix} \times 2$\\

     \midrule
     conv4\_x & $3\times3\times3$ max pool, stride 2 \\
      & $\begin{bmatrix}
            3\times3\times3, & 256\\ 
            3\times3\times3, & 256 
            \end{bmatrix} \times 2$\\
     \midrule
     conv3\_x & $3\times3\times3$ max pool, stride 2 \\
      & $\begin{bmatrix}
            3\times3\times3, & 512\\ 
            3\times3\times3, & 512 
            \end{bmatrix} \times 2$\\
     \midrule
    &average pool\\
    \midrule
    \midrule
    projection head & $512\times2048$ FC \\
    & BatchNorm1d \\
    & ReLU \\
    &$2048\times128$ FC \\
    \midrule 
    linear layer & $512 \times \text{number of classes}$ FC\\
    \bottomrule
    \end{tabular}%
    }
\vspace{-0.4cm}
\end{table}
\begin{algorithm}
\SetAlgoLined
\textbf{Input:} encoder $f_\theta(\cdot)$, unlabeled data $X$
\\
\textbf{Procedure:}\\
 \While{not max epoch}{
    \If{epoch \% cluster interval = 0}{
        $Z$ = $f_\theta(X)$\\
        $\{P_i\}$ = FINCH($Z$)\\
         select cluster assignments $C$ from partition $P_1$. $C=\{1,2,..., C_{P_1}\}$
    }
    
    \For{x in DataLoader(X)}{
        $z$ = $f_\theta(x)$ \\
        $z^+$ = PositiveMining($z$, $C$)\\
        $z^-$ = NegativeMining($z$, $z^+$, $C$) \\
        $\mathcal{L}_{triplet}$ = TripletMarginLoss($z, z^+, z^-; \theta, m_1$)\\
        $\mathcal{L}_{temporal}$  = TripletMarginLoss($z$, \texttt{aug}$(z)$, $z^+$; $\theta, m_2$)\\
        $\mathcal{L}_{total}$  = $\mathcal{L}_{triplet}$  + $\lambda \mathcal{L}_{temporal}$\\ 
        $\theta$ = SGD($\mathcal{L}_{total}$, $\theta$)\\
        }
 }
\caption{SLIC Training Algorithm}
\label{alg:training}
\end{algorithm}

\section{Training Algorithm Overview}
See Algorithm \ref{alg:training} for details.
\label{algorithm_overview}


\section{Pretraining using the InfoNCE loss}
\label{slic_infonce}
To examine how performance would vary if a different loss function is used for instance-based discrimination, we pretrain the model using the InfoNCE loss \cite{pmlr-v9-gutmann10a} (replacing our instance-based triplet loss) with a batch size of 32 across 2 GPUs, and temperature set to 0.1 (these results are not meant to be an improved version of SLIC, they are a more detailed analysis on the loss function used). The results for action recognition and video retrieval are presented in Table \ref{tbl:infoNCE_finetune_res} and \ref{tbl:infoNCE_NN_retrieval} (the SLIC results are those presented in Section \ref{experiment}). When using the InfoNCE loss during pretraining, SLIC outperforms its triplet loss counterpart in top-1 video retrieval for UCF101, but underperforms in all other evaluation tasks. 

\begin{table}[h!]
  \caption{\textbf{Linear probing and end-to-end finetuning results for action classification using the InfoNCE loss} pretrained on UCF101, then fine-tuned on UCF101 and HMDB51, using only visual inputs. The right-most columns indicate the top-1 accuracy for each dataset. `Frozen \cmark' indicates classification with a linear layer on top of a frozen backbone; `Frozen \xmark' indicates end-to-end finetuning.}
  \label{tbl:infoNCE_finetune_res}
  \centering
  \resizebox{\columnwidth}{!}{%
  \begin{tabu}{lccccccc}
    \toprule
    Method & Input Size & Arch. & Frozen & UCF101 & HMDB51 \\
    \midrule
    SLIC  & $16 \times 128^2$ & R3D & \cmark & 72.3 & 41.8	 \\
  \textbf{SLIC} & $32 \times 128^2$ & R3D & \cmark & \textbf{77.7} & \textbf{48.3} 	 \\
  SLIC-infoNCE    & $16 \times 128^2$ & R3D & \cmark & 70.2 & 39.8 	 \\ 
  SLIC-infoNCE  & $32 \times 128^2$&  R3D & \cmark & 75.4 & 44.7 	 \\ 
    \midrule
    \midrule

    SLIC  & $16 \times 128^2$ & R3D & \xmark & 77.4 & 46.2 \\
    \textbf{SLIC}   & $32 \times 128^2$ & R3D & \xmark & \textbf{83.2} & \textbf{54.5}  \\ 
    SLIC-InfoNCE  & $16 \times 128^2$ & R3D & \xmark & 76.8 & 45.5 \\
    SLIC-InfoNCE   & $32 \times 128^2$ & R3D& \xmark & 81.9 & 53.6 \\

    \bottomrule

  \end{tabu}%
  }
  \vspace{-0.2cm}
\end{table}



    

\begin{table}[ht]
  \caption{\textbf{Nearest neighbour video retrieval results using the InfoNCE loss} on UCF101 and HMDB51 (both split-1). Testing set clips are used as queries to retrieve the top-$k$ nearest neighbors in the training set, where $k \in [1,5,10,20]$. $N_T$ indicates the temporal input size.}
  \label{tbl:infoNCE_NN_retrieval}
  \centering
  \resizebox{\columnwidth}{!}{
  \begin{tabu}{lcc|c|c|c}
    \toprule
    \multicolumn{2}{c}{}    &
    \multicolumn{4}{c}{UCF101 / HMDB51} \\
    \cmidrule(r){3-6}
    Method & $N_T$ &  R@1 & R@5 & R@10 & R@20 \\
    \midrule
    SLIC & 16 &  66.7 / 25.3 & 77.3 / 49.8 & 82.0 / 64.9 & 86.4 / 76.1 \\
    SLIC & 32  &  71.6 / \textbf{28.9} & \textbf{82.4} / \textbf{52.8} & \textbf{86.6} / \textbf{65.4} & \textbf{90.3} / \textbf{77.8} \\
     SLIC-InfoNCE & 16 &  66.8 / 21.4 & 74.5 / 45.1 & 77.7 / 57.2 & 81.3 / 71.6 \\
     SLIC-InfoNCE & 32 &  \textbf{74.4} / 26.2 & 81.0 / 49.7 & 84.1 / 62.5 & 87.1 / 75.9 \\
    
    \bottomrule
  \end{tabu}
  }
\end{table}

    

\section{Retrieval Results without Projection Head}
\label{retrieval_results_wo_projection_head}
In this section, we present the nearest neighbour retrieval results on UCF101 and HMDB51 without the projection head. After pretraining on the UCF101 dataset, we discard the non-linear projection head and use the intermediate embeddings for retrieval evaluation. The results are presented in Table~\ref{tbl:NN_retrieval_no_ph}. When evaluating without the non-linear projection head on retrieval, SLIC with 32-frame input size achieved similar results compared to Table~\ref{tbl:NN_retrieval}.
\begin{table}[ht]
  \vspace{-0.15cm}
  \caption{\textbf{
  Video retrieval results without projection head} on UCF101 and HMDB51.
  $N_T$ indicates the temporal input size. } 
  \vspace{-0.2cm}
  \label{tbl:NN_retrieval_no_ph}
  \centering
  \resizebox{\columnwidth}{!}{%
  \begin{tabu}{lcc|c|c|c}
    \toprule
    \multicolumn{2}{c}{}    & \multicolumn{4}{c}{UCF101 / HMDB51} \\  
    \cmidrule(r){3-6}
    Method & $N_T$ & R@1 & R@5 & R@10 & R@20 \\
    \midrule
    CoCLR-RGB\cite{coclr}   & 32 & 53.3 / 23.2 & 69.4 / 43.2 & 76.6 / 53.5 & 82.0 / 65.5 \\
    TCLR\cite{dave2021tclr} & 16 & 56.2 / 22.8 & 72.2 / 45.4 & 79.0 / 57.8 & 85.3 / 73.1 \\
    \midrule
    SLIC (S3D-23) & 16 & 58.9 / 22.4 & 74.0 / 47.3 & 80.7 / 60.6 & 86.2 / 74.1 \\
    SLIC (S3D-23) & 32 & 67.5 / 28.0 & 80.1 / 55.1 & 85.2 / 67.6 & 90.0 / 79.6 \\
    SLIC (R3D-18) & 16 & 61.0 / 24.4 & 76.2 / 49.0 & 83.1 / 62.6 & 88.8 / 75.2 \\
    SLIC (R3D-18) & 32 & \textbf{70.8} / \textbf{28.9} & \textbf{84.6} / \textbf{55.6} & \textbf{89.0} / \textbf{68.8} & \textbf{92.7} / \textbf{81.3} \\
    \bottomrule
  \end{tabu}%
  }

\end{table}

\section{Classification using RGB and optical flow}
\label{action_classif_dualview}

We additionally conduct a study on the effect of using both the RGB and optical flow views during finetuning for action classification. The results from this study are not meant to be an improved version of SLIC, instead they are only meant to show that using both views can improve performance at the cost of additional memory (from a separate optical flow classification model) and processing time (from computing optical flow for test clips). After pretraining on Kinetics,
we finetune the same pretrained model on RGB and optical flow inputs separately, 
then average the predictions. We compare the results to other methods that use 2-stream inputs, and present the results in Table \ref{tbl:multiview_res}. It is observed in all methods that using multi-view information during finetuning helps improve the action classification results on both UCF101 and HMDB51. 

\begin{table}[h!]
  \caption{\textbf{End-to-end finetuning top-1 accuracy results for action classification} pretrained on Kinetics400, then fine-tuned on UCF101 and HMDB51, using only visual inputs.
  \color{gray}
  {Methods with \textsuperscript{\textdagger} use optical flow in addition to RGB as inputs to the action classification model.}
  \color{black}{SLIC\textsuperscript{\textasciicircum} indicates that pretraining used 16 frame inputs. Dashes indicate unavailable information.}}
  \label{tbl:multiview_res}
    \vspace{-0.1cm}

  \centering
  
  \resizebox{\columnwidth}{!}{%
  \begin{tabu}{lcccc}
    \toprule
    Pretrain. Method & Input Size & Arch. & UCF101 & HMDB51 \\
    \toprule
    
    \midrule
    CoCon-RGB\textsuperscript{*}\cite{CoCon2021} & $(-)\times 224^{2}$ & R3D-18 & 71.6 & 46.0 \\
    \rowfont{\color{gray}}
    CoCon-E\textsuperscript{*}\textsuperscript{\textdagger}\cite{CoCon2021} & $(-)\times 224^{2}$ & R3D-18 & 78.1 & 52.0 \\
    \textbf{CoCLR-RGB\cite{coclr}} & $32 \times 128^2$ & S3D-23 & 87.9 & 54.6 \\
    \rowfont{\color{gray}}
    CoCLR\textsuperscript{\textdagger}\cite{coclr} & $32 \times 128^2$ & S3D-23 & \textbf{90.6} & \textbf{62.9} \\
    \midrule
    SLIC\textsuperscript{\textasciicircum} & $32 \times 128^2$ & R3D-18  & 83.2 & 52.2 \\
    \rowfont{\color{gray}}
    SLIC\textsuperscript{\textasciicircum}\textsuperscript{\textdagger} & $32 \times 128^2$ & R3D-18  & 86.3 & 57.3\\

    \bottomrule

  \end{tabu}%
  }
  \vspace{-0.20cm}
\end{table}

\section{Visualization of learned representations}
\label{tsne_visualization}
\begin{figure*}[ht]
  \centering
  \includegraphics[width=0.93\textwidth]{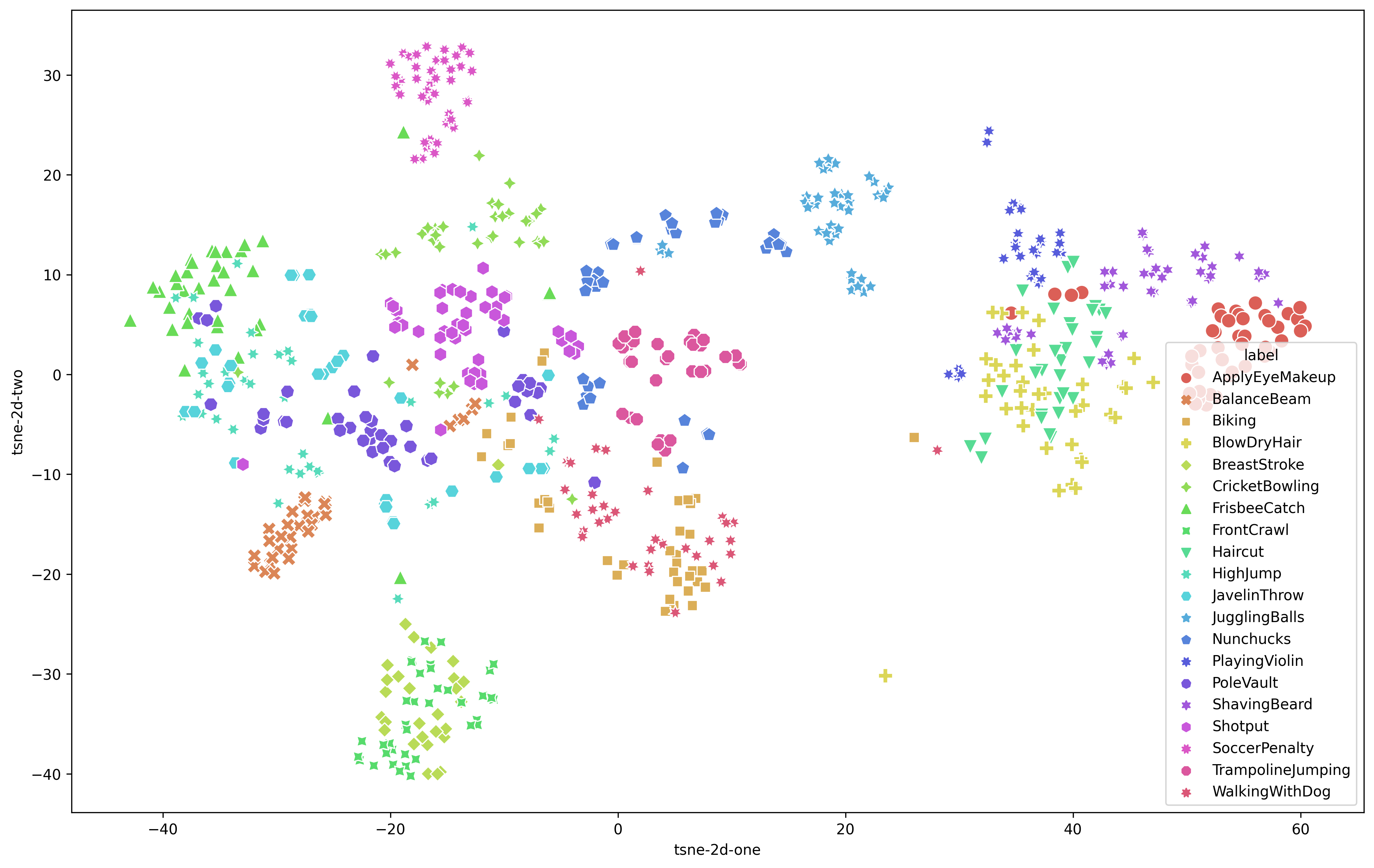}
  \caption{t-SNE visualization of SLIC embeddings for 20 randomly chosen action classes from the UCF101 test set (split-1).}
  \label{fig:tsne_slic_val}
\end{figure*}

\begin{figure*}[ht]
  \centering
  \includegraphics[width=0.93\textwidth]{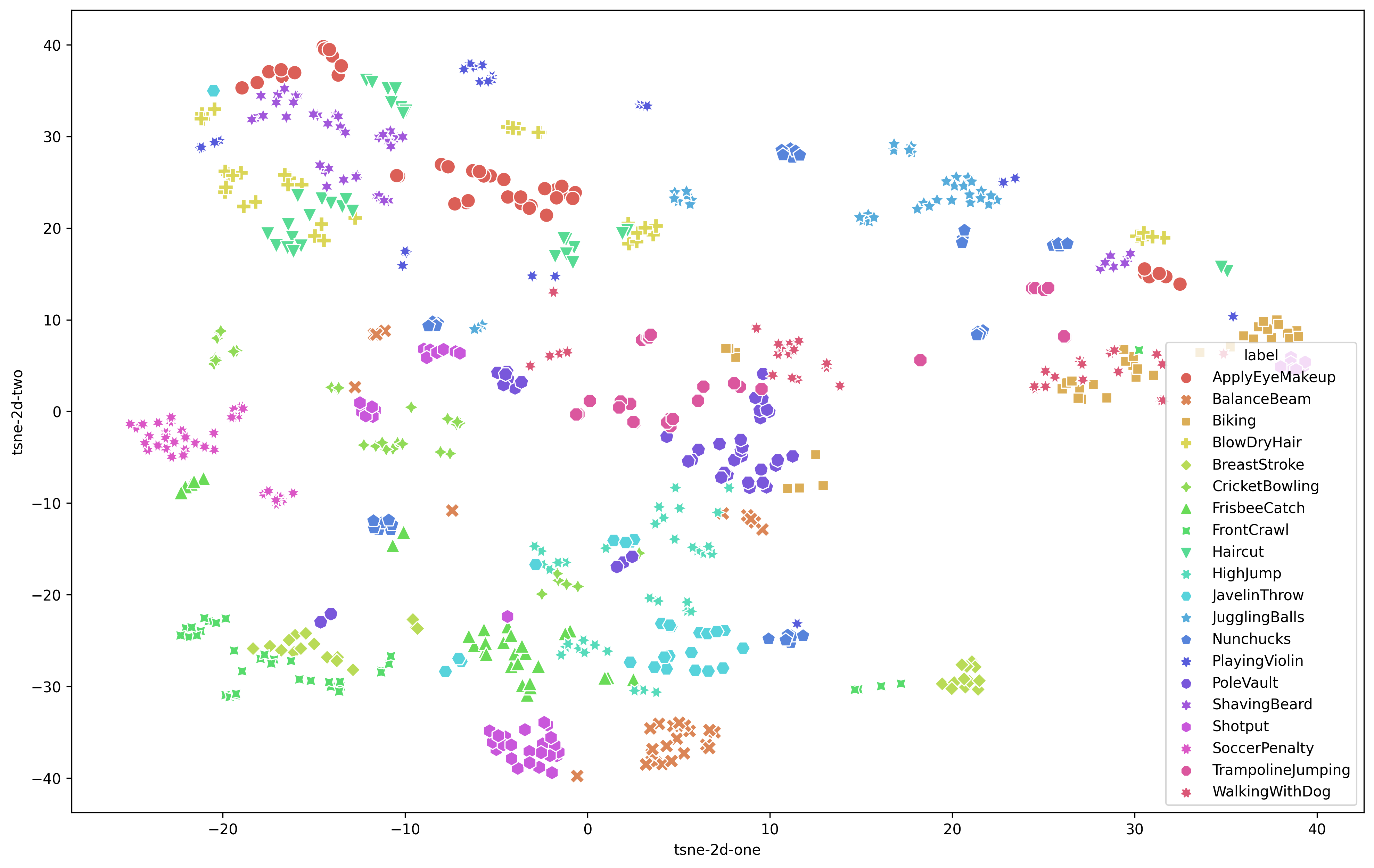}
  \caption{t-SNE visualization of CoCLR embeddings for 20 randomly chosen action classes from the UCF101 test set (split-1).}
  \label{fig:tsne_coclr_val}
\end{figure*}

We use t-SNE to visualize the learned representations for the UCF101 test set from self-supervised pretraining on the UCF101 training set (split-1). We compare our method (the version trained with 16 frame inputs) to CoCLR\cite{coclr} for 20 randomly selected classes (for better visibility) from UCF101. For SLIC, the center 16 frames are used as inputs, and for CoCLR, the center 32 frames are used as inputs since CoCLR was trained using 32-frame inputs. In both cases, the embeddings visualized are those that were used for the video retrieval task (the 128-d outputs of the non-linear projection head for SLIC, and the 1024-d backbone features for CoCLR), and they are reduced to 50 dimensions using PCA. The perplexity for t-SNE is set to 30. By comparing Figure \ref{fig:tsne_slic_val} and Figure \ref{fig:tsne_coclr_val}, we can see that SLIC representations result in well separated clusters compared to those of CoCLR.

\section{Confusion matrix for action classification}
\label{confusion_matrix_classification}

\begin{figure*}[ht]
  \centering
  \includegraphics[width=0.93\textwidth]{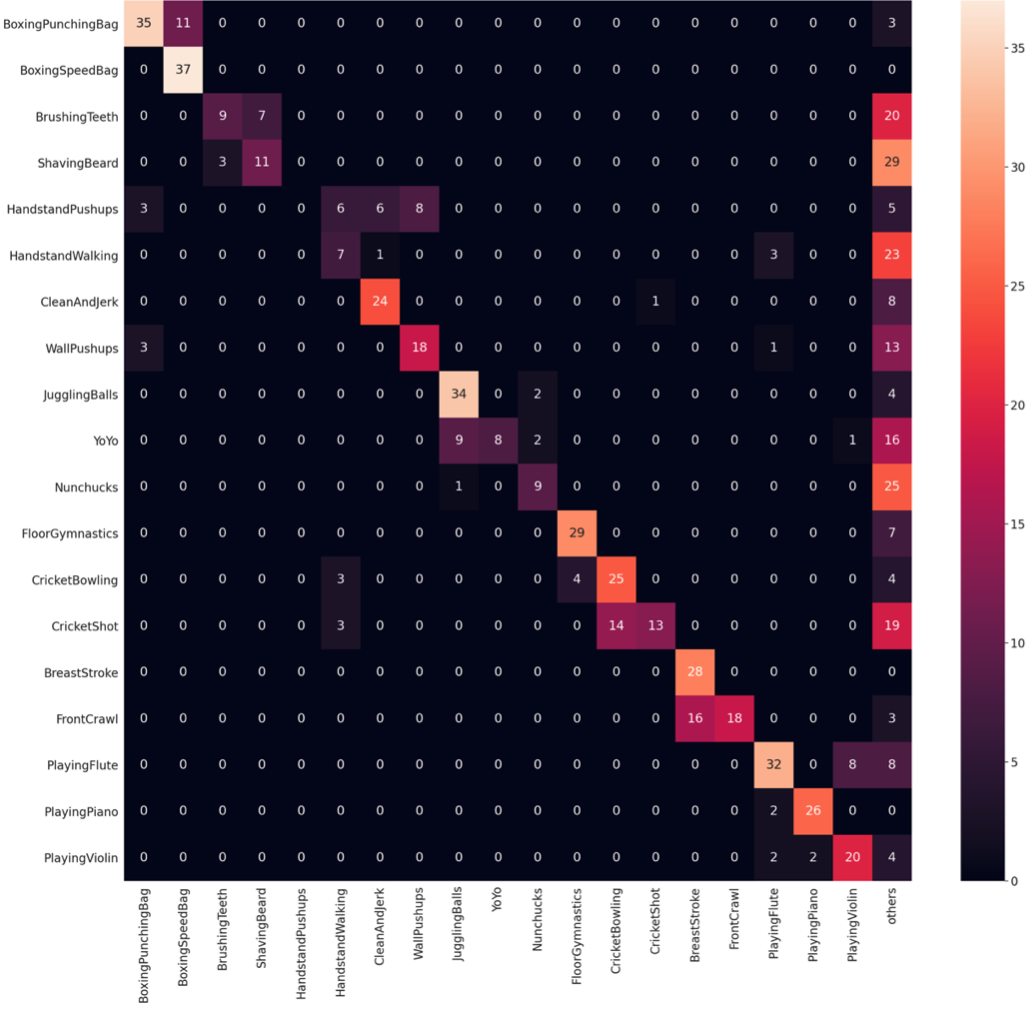}
  \caption{Confusion matrix of UCF101 generated by SLIC after pretraining on UCF101.}
  \label{fig:confusion_matrix}
\end{figure*}

\begin{figure*}[ht]
  \centering
  \includegraphics[width=0.93\textwidth]{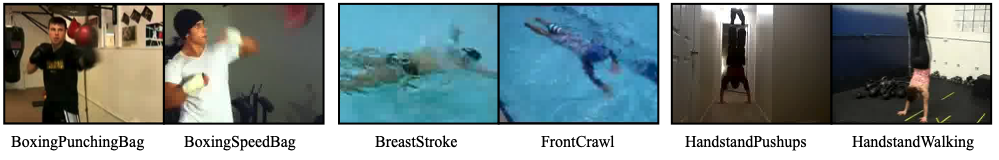}
  \caption{Visualization of 3 pairs of action classes that are easily confused.}
  \label{fig:confused_classes}
\end{figure*}

We visualize the confusion matrix of the 19 most confusing classes from UCF101 after end-to-end finetuning in Figure \ref{fig:confusion_matrix}. We can see that most action class pairs that are poorly distinguished are likely difficult due to their visually similar semantics. Examples of these confusing action class pairs are visualized in Figure \ref{fig:confused_classes}. 

%% file: PaperForReview.bbl
\begin{thebibliography}{10}\itemsep=-1pt

\bibitem{akbari2021vatt}
Hassan Akbari, Linagzhe Yuan, Rui Qian, Wei-Hong Chuang, Shih-Fu Chang, Yin
  Cui, and Boqing Gong.
\newblock Vatt: Transformers for multimodal self-supervised learning from raw
  video, audio and text, 2021.

\bibitem{alwassel2020selfsupervised}
Humam Alwassel, Dhruv Mahajan, Bruno Korbar, Lorenzo Torresani, Bernard Ghanem,
  and Du Tran.
\newblock Self-supervised learning by cross-modal audio-video clustering, 2020.

\bibitem{cross_modal}
Humam Alwassel, Dhruv Mahajan, Lorenzo Torresani, Bernard Ghanem, and Du Tran.
\newblock Self-supervised learning by cross-modal audio-video clustering.
\newblock {\em CoRR}, abs/1911.12667, 2019.

\bibitem{arandjelovic2017look}
Relja Arandjelović and Andrew Zisserman.
\newblock Look, listen and learn.
\newblock {\em arXiv:1705.08168}, 2017.

\bibitem{label_multimodal}
Yuki~Markus Asano, Mandela Patrick, Christian Rupprecht, and Andrea Vedaldi.
\newblock Labelling unlabelled videos from scratch with multi-modal
  self-supervision.
\newblock {\em CoRR}, abs/2006.13662, 2020.

\bibitem{Benaim_2020_CVPR}
Sagie Benaim, Ariel Ephrat, Oran Lang, Inbar Mosseri, William~T. Freeman,
  Michael Rubinstein, Michal Irani, and Tali Dekel.
\newblock Speednet: Learning the speediness in videos.
\newblock In {\em Proceedings of the IEEE/CVF Conference on Computer Vision and
  Pattern Recognition (CVPR)}, June 2020.

\bibitem{cai2020negatives}
Tiffany~Tianhui Cai, Jonathan Frankle, David~J. Schwab, and Ari~S. Morcos.
\newblock Are all negatives created equal in contrastive instance
  discrimination?, 2020.

\bibitem{caron2018deep}
Mathilde Caron, Piotr Bojanowski, Armand Joulin, and Matthijs Douze.
\newblock Deep clustering for unsupervised learning of visual features.
\newblock In {\em European Conference on Computer Vision}, 2018.

\bibitem{caron2019unsupervised}
Mathilde Caron, Piotr Bojanowski, Julien Mairal, and Armand Joulin.
\newblock Unsupervised pre-training of image features on non-curated data.
\newblock {\em arXiv:1905.01278}, 2019.

\bibitem{swav}
Mathilde Caron, Ishan Misra, Julien Mairal, Priya Goyal, Piotr Bojanowski, and
  Armand Joulin.
\newblock Unsupervised learning of visual features by contrasting cluster
  assignments.
\newblock In {\em Advances in Neural Information Processing Systems},
  volume~33, pages 9912--9924. Curran Associates, Inc., 2020.

\bibitem{carreira2017i3d}
João Carreira and Andrew Zisserman.
\newblock Quo vadis, action recognition? a new model and the kinetics dataset.
\newblock In {\em 2017 IEEE Conference on Computer Vision and Pattern
  Recognition (CVPR)}, pages 4724--4733, 2017.

\bibitem{chen2020RSPNet}
Peihao Chen, Deng Huang, Dongliang He, Xiang Long, Runhao Zeng, Shilei Wen,
  Mingkui Tan, and Chuang Gan.
\newblock Rspnet: Relative speed perception for unsupervised video
  representation learning.
\newblock In {\em The AAAI Conference on Artificial Intelligence (AAAI)}, 2021.

\bibitem{chen2020simple}
Ting Chen, Simon Kornblith, Mohammad Norouzi, and Geoffrey Hinton.
\newblock A simple framework for contrastive learning of visual
  representations.
\newblock {\em arXiv preprint arXiv:2002.05709}, 2020.

\bibitem{chen2020mocov2}
Xinlei Chen, Haoqi Fan, Ross Girshick, and Kaiming He.
\newblock Improved baselines with momentum contrastive learning.
\newblock {\em arXiv preprint arXiv:2003.04297}, 2020.

\bibitem{dave2021tclr}
Ishan Dave, Rohit Gupta, Mamshad~Nayeem Rizve, and Mubarak Shah.
\newblock Tclr: Temporal contrastive learning for video representation.
\newblock {\em arXiv:2101.07974}, 2021.

\bibitem{EM_algo}
A.~P. Dempster, N.~M. Laird, and D.~B. Rubin.
\newblock Maximum likelihood from incomplete data via the em algorithm.
\newblock {\em Journal of the Royal Statistical Society. Series B
  (Methodological)}, 39(1):1--38, 1977.

\bibitem{dynamonet}
Ali Diba, Vivek Sharma, Luc Van~Gool, and Rainer Stiefelhagen.
\newblock Dynamonet: Dynamic action and motion network.
\newblock In {\em 2019 IEEE/CVF International Conference on Computer Vision
  (ICCV)}, pages 6191--6200, 2019.

\bibitem{doersch2015unsupervised}
Carl Doersch, Abhinav Gupta, and Alexei~A. Efros.
\newblock Unsupervised visual representation learning by context prediction.
\newblock In {\em International Conference on Computer Vision (ICCV)}, 2015.

\bibitem{8237488}
Carl Doersch and Andrew Zisserman.
\newblock Multi-task self-supervised visual learning.
\newblock In {\em 2017 IEEE International Conference on Computer Vision
  (ICCV)}, pages 2070--2079, 2017.

\bibitem{dbscan}
Martin Ester, Hans-Peter Kriegel, J\"{o}rg Sander, and Xiaowei Xu.
\newblock A density-based algorithm for discovering clusters in large spatial
  databases with noise.
\newblock In {\em International Conference on Knowledge Discovery and Data
  Mining}, KDD'96, page 226–231. AAAI Press, 1996.

\bibitem{Feichtenhofer_2020_CVPR}
Christoph Feichtenhofer.
\newblock X3d: Expanding architectures for efficient video recognition.
\newblock In {\em Computer Vision and Pattern Recognition (CVPR)}, June 2020.

\bibitem{feichtenhofer2019slowfast}
Christoph Feichtenhofer, Haoqi Fan, Jitendra Malik, and Kaiming He.
\newblock Slowfast networks for video recognition, 2019.

\bibitem{feichtenhofer2016convolutional}
Christoph Feichtenhofer, Axel Pinz, and Andrew Zisserman.
\newblock Convolutional two-stream network fusion for video action recognition.
\newblock In {\em Conference on Computer Vision and Pattern Recognition
  (CVPR)}, 2016.

\bibitem{gidaris2018unsupervised}
Spyros Gidaris, Praveer Singh, and Nikos Komodakis.
\newblock Unsupervised representation learning by predicting image rotations.
\newblock In {\em International Conference on Learning Representations}, 2018.

\bibitem{byol}
Jean-Bastien Grill, Florian Strub, Florent Altch\'{e}, Corentin Tallec, Pierre
  Richemond, Elena Buchatskaya, Carl Doersch, Bernardo Avila~Pires, Zhaohan
  Guo, Mohammad Gheshlaghi~Azar, Bilal Piot, koray kavukcuoglu, Remi Munos, and
  Michal Valko.
\newblock Bootstrap your own latent - a new approach to self-supervised
  learning.
\newblock In {\em Advances in Neural Information Processing Systems},
  volume~33, pages 21271--21284. Curran Associates, Inc., 2020.

\bibitem{pmlr-v9-gutmann10a}
Michael Gutmann and Aapo Hyvärinen.
\newblock Noise-contrastive estimation: A new estimation principle for
  unnormalized statistical models.
\newblock In Yee~Whye Teh and Mike Titterington, editors, {\em International
  Conference on Artificial Intelligence and Statistics}, volume~9 of {\em
  Proceedings of Machine Learning Research}, pages 297--304, Chia Laguna
  Resort, Sardinia, Italy, 13--15 May 2010. PMLR.

\bibitem{DPC}
Tengda Han, Weidi Xie, and Andrew Zisserman.
\newblock Video representation learning by dense predictive coding.
\newblock {\em CoRR}, abs/1909.04656, 2019.

\bibitem{Han20}
Tengda Han, Weidi Xie, and Andrew Zisserman.
\newblock Memory-augmented dense predictive coding for video representation
  learning.
\newblock In {\em European Conference on Computer Vision}, 2020.

\bibitem{coclr}
Tengda Han, Weidi Xie, and Andrew Zisserman.
\newblock Self-supervised co-training for video representation learning.
\newblock In H. Larochelle, M. Ranzato, R. Hadsell, M.~F. Balcan, and H. Lin,
  editors, {\em Advances in Neural Information Processing Systems}, volume~33,
  pages 5679--5690. Curran Associates, Inc., 2020.

\bibitem{hara2018spatiotemporal}
Kensho Hara, Hirokatsu Kataoka, and Yutaka Satoh.
\newblock Can spatiotemporal 3d cnns retrace the history of 2d cnns and
  imagenet?, 2018.

\bibitem{3d_conv_spatiotemporal}
Kensho Hara, Hirokatsu Kataoka, and Yutaka Satoh.
\newblock Towards good practice for action recognition with spatiotemporal 3d
  convolutions.
\newblock In {\em 24th International Conference on Pattern Recognition (ICPR)},
  pages 2516--2521, 2018.

\bibitem{VideoDeepInfoMax}
R.~Devon Hjelm and Philip Bachman.
\newblock Representation learning with video deep infomax.
\newblock {\em CoRR}, abs/2007.13278, 2020.

\bibitem{hsu2016neural}
Yen-Chang Hsu and Zsolt Kira.
\newblock Neural network-based clustering using pairwise constraints.
\newblock {\em arXiv:1511.06321}, 2016.

\bibitem{kay2017kinetics}
Will Kay, Joao Carreira, Karen Simonyan, Brian Zhang, Chloe Hillier, Sudheendra
  Vijayanarasimhan, Fabio Viola, Tim Green, Trevor Back, Paul Natsev, Mustafa
  Suleyman, and Andrew Zisserman.
\newblock The kinetics human action video dataset.
\newblock {\em arXiv:1705.06950}, 2017.

\bibitem{supcon}
Prannay Khosla, Piotr Teterwak, Chen Wang, Aaron Sarna, Yonglong Tian, Phillip
  Isola, Aaron Maschinot, Ce Liu, and Dilip Krishnan.
\newblock Supervised contrastive learning.
\newblock In H. Larochelle, M. Ranzato, R. Hadsell, M.~F. Balcan, and H. Lin,
  editors, {\em Advances in Neural Information Processing Systems}, volume~33,
  pages 18661--18673. Curran Associates, Inc., 2020.

\bibitem{korbar}
Bruno Korbar, Du Tran, and Lorenzo Torresani.
\newblock Cooperative learning of audio and video models from self-supervised
  synchronization.
\newblock In {\em Proceedings of the 32nd International Conference on Neural
  Information Processing Systems}, page 7774–7785, 2018.

\bibitem{hmdb}
H. Kuehne, H. Jhuang, E. Garrote, T. Poggio, and T. Serre.
\newblock Hmdb: A large video database for human motion recognition.
\newblock In {\em 2011 International Conference on Computer Vision}, pages
  2556--2563, 2011.

\bibitem{PCL}
Junnan Li, Pan Zhou, Caiming Xiong, and Steven~C.H. Hoi.
\newblock Prototypical contrastive learning of unsupervised representations.
\newblock {\em ICLR}, 2021.

\bibitem{VCP}
Dezhao Luo, Chang Liu, Yu Zhou, Dongbao Yang, Can Ma, Qixiang Ye, and Weiping
  Wang.
\newblock Video cloze procedure for self-supervised spatio-temporal learning,
  2020.

\bibitem{finch}
Vivek~Sharma M.~Saquib~Sarfraz and Rainer Stiefelhagen.
\newblock Efficient parameter-free clustering using first neighbor relations.
\newblock In {\em Computer Vision and Pattern Recognition (CVPR)}, pages
  8934--8943, 2019.

\bibitem{Noroozi2016UnsupervisedLO}
M. Noroozi and P. Favaro.
\newblock Unsupervised learning of visual representations by solving jigsaw
  puzzles.
\newblock In {\em ECCV}, 2016.

\bibitem{videoMoCo}
Tian Pan, Yibing Song, Tianyu Yang, Wenhao Jiang, and Wei Liu.
\newblock Videomoco: Contrastive video representation learning with temporally
  adversarial examples.
\newblock {\em CoRR}, abs/2103.05905, 2021.

\bibitem{pathakCVPR16context}
Deepak Pathak, Philipp Kr\"ahenb\"uhl, Jeff Donahue, Trevor Darrell, and Alexei
  Efros.
\newblock Context encoders: Feature learning by inpainting.
\newblock In {\em CVPR}, 2016.

\bibitem{qian2021spatiotemporal}
Rui Qian, Tianjian Meng, Boqing Gong, Ming-Hsuan Yang, Huisheng Wang, Serge
  Belongie, and Yin Cui.
\newblock Spatiotemporal contrastive video representation learning.
\newblock {\em arXiv:2008.03800}, 2021.

\bibitem{CLIP}
Alec Radford, Jong~Wook Kim, Chris Hallacy, Aditya Ramesh, Gabriel Goh,
  Sandhini Agarwal, Girish Sastry, Amanda Askell, Pamela Mishkin, Jack Clark,
  Gretchen Krueger, and Ilya Sutskever.
\newblock Learning transferable visual models from natural language
  supervision, 2021.

\bibitem{CoCon2021}
Nishant Rai, Ehsan Adeli, Kuan-Hui Lee, Adrien Gaidon, and Juan~Carlos Niebles.
\newblock Cocon: Cooperative-contrastive learning.
\newblock In {\em Proceedings of the IEEE/CVF Conference on Computer Vision and
  Pattern Recognition (CVPR) Workshops}, pages 3384--3393, June 2021.

\bibitem{Schroff_2015}
Florian Schroff, Dmitry Kalenichenko, and James Philbin.
\newblock Facenet: A unified embedding for face recognition and clustering.
\newblock {\em Conference on Computer Vision and Pattern Recognition (CVPR)},
  Jun 2015.

\bibitem{TimeContrastiveNetworks}
Pierre Sermanet, Corey Lynch, Jasmine Hsu, and Sergey Levine.
\newblock Time-contrastive networks: Self-supervised learning from multi-view
  observation.
\newblock {\em CoRR}, abs/1704.06888, 2017.

\bibitem{face_finch}
Vivek Sharma, Makarand Tapaswi, M.~Saquib Sarfraz, and Rainer Stiefelhagen.
\newblock Clustering based contrastive learning for improving face
  representations.
\newblock In {\em 2020 15th IEEE International Conference on Automatic Face and
  Gesture Recognition (FG 2020)}, pages 109--116, 2020.

\bibitem{simonyan}
Karen Simonyan and Andrew Zisserman.
\newblock Two-stream convolutional networks for action recognition in videos.
\newblock In {\em International Conference on Neural Information Processing
  Systems - Volume 1}, NIPS'14, page 568–576, Cambridge, MA, USA, 2014. MIT
  Press.

\bibitem{npairs}
Kihyuk Sohn.
\newblock Improved deep metric learning with multi-class n-pair loss objective.
\newblock In D. Lee, M. Sugiyama, U. Luxburg, I. Guyon, and R. Garnett,
  editors, {\em Advances in Neural Information Processing Systems}, volume~29.
  Curran Associates, Inc., 2016.

\bibitem{soomro2012ucf101}
Khurram Soomro, Amir~Roshan Zamir, and Mubarak Shah.
\newblock Ucf101: A dataset of 101 human actions classes from videos in the
  wild.
\newblock {\em arXiv:1212.0402}, 2012.

\bibitem{iic}
Li Tao, Xueting Wang, and Toshihiko Yamasaki.
\newblock Self-supervised video representation learning using inter-intra
  contrastive framework.
\newblock {\em Proceedings of the 28th ACM International Conference on
  Multimedia}, Oct 2020.

\bibitem{tran2018closer}
Du Tran, Heng Wang, Lorenzo Torresani, Jamie Ray, Yann LeCun, and Manohar
  Paluri.
\newblock A closer look at spatiotemporal convolutions for action recognition.
\newblock {\em arXiv:1711.11248}, 2018.

\bibitem{oord2019representation}
Aaron van~den Oord, Yazhe Li, and Oriol Vinyals.
\newblock Representation learning with contrastive predictive coding.
\newblock {\em arXiv:1807.03748}, 2019.

\bibitem{vondrick2018tracking}
Carl Vondrick, Abhinav Shrivastava, Alireza Fathi, Sergio Guadarrama, and Kevin
  Murphy.
\newblock Tracking emerges by colorizing videos.
\newblock {\em arXiv:1806.09594}, 2018.

\bibitem{spatio_tempoal_statistics}
Jiangliu Wang, Jianbo Jiao, Linchao Bao, Shengfeng He, Wei Liu, and Yunhui Liu.
\newblock Self-supervised video representation learning by uncovering
  spatio-temporal statistics.
\newblock {\em CoRR}, abs/2008.13426, 2020.

\bibitem{Wang20}
Jiangliu Wang, Jianbo Jiao, and Yunhui Liu.
\newblock Self-supervised video representation learning by pace prediction.
\newblock In {\em European Conference on Computer Vision}, 2020.

\bibitem{wei2018learning}
Donglai Wei, Joseph~J Lim, Andrew Zisserman, and William~T Freeman.
\newblock Learning and using the arrow of time.
\newblock In {\em Computer Vision and Pattern Recognition (CVPR)}, pages
  8052--8060, 2018.

\bibitem{xie2018rethinking}
Saining Xie, Chen Sun, Jonathan Huang, Zhuowen Tu, and Kevin Murphy.
\newblock Rethinking spatiotemporal feature learning: Speed-accuracy trade-offs
  in video classification.
\newblock {\em arXiv:1712.04851}, 2018.

\bibitem{VCOP}
Dejing Xu, Jun Xiao, Zhou Zhao, Jian Shao, Di Xie, and Yueting Zhuang.
\newblock Self-supervised spatiotemporal learning via video clip order
  prediction.
\newblock In {\em Computer Vision and Pattern Recognition (CVPR)}, June 2019.

\bibitem{yan2019clusterfit}
Xueting Yan, Ishan Misra, Abhinav Gupta, Deepti Ghadiyaram, and Dhruv Mahajan.
\newblock Clusterfit: Improving generalization of visual representations.
\newblock {\em arXiv:1912.03330}, 2019.

\bibitem{SeCo2021}
Ting Yao, Yiheng Zhang, Zhaofan Qiu, Yingwei Pan, and Tao Mei.
\newblock Seco: Exploring sequence supervision for unsupervised representation
  learning.
\newblock In {\em 35th AAAI Conference on Artificial Intelligence}, 2021.

\bibitem{yao2020video}
Yuan Yao, Chang Liu, Dezhao Luo, Yu Zhou, and Qixiang Ye.
\newblock Video playback rate perception for self-supervised spatio-temporal
  representation learning.
\newblock {\em arXiv:2006.11476}, 2020.

\bibitem{YM.2020Self-labelling}
Asano Y.M., Rupprecht C., and Vedaldi A.
\newblock Self-labelling via simultaneous clustering and representation
  learning.
\newblock In {\em International Conference on Learning Representations}, 2020.

\bibitem{zach2007tvl1}
C. Zach, T. Pock, and H. Bischof.
\newblock A duality based approach for realtime tv-l1 optical flow.
\newblock In {\em Proceedings of the 29th DAGM Conference on Pattern
  Recognition}, page 214–223, Berlin, Heidelberg, 2007. Springer-Verlag.

\bibitem{zhang2016colorful}
Richard Zhang, Phillip Isola, and Alexei~A Efros.
\newblock Colorful image colorization.
\newblock In {\em ECCV}, 2016.

\bibitem{Zhao_2019_CVPR}
Jiaojiao Zhao and Cees G.~M. Snoek.
\newblock Dance with flow: Two-in-one stream action detection.
\newblock In {\em Proceedings of the IEEE/CVF Conference on Computer Vision and
  Pattern Recognition (CVPR)}, June 2019.

\end{thebibliography}
